\pdfoutput=1

\documentclass[11pt, hyphens]{article}
\usepackage{tcolorbox}

\usepackage[final]{acl}

\usepackage{times}
\usepackage{latexsym}
\usepackage{amsfonts}

\usepackage[T1]{fontenc}
\usepackage{amsmath}
\usepackage[utf8]{inputenc}
\usepackage{enumitem}
\usepackage{microtype}

\usepackage{inconsolata}

\usepackage{graphicx}
\usepackage{cleveref}
\usepackage{fontawesome}

\usepackage{multirow}
\usepackage{multicol}
\usepackage{tabularx}
\usepackage{booktabs}
\usepackage{xcolor}
\usepackage{adjustbox}
\usepackage{makecell}

\crefformat{section}{\S#2#1#3} 
\crefformat{subsection}{\S#2#1#3}
\crefformat{subsubsection}{\S#2#1#3}

\setlist{noitemsep, leftmargin=*, topsep=0pt, partopsep=0pt}

\newcommand{\SYSTEM}{FUEL}

\newcommand*{\Scale}[2][4]{\scalebox{#1}{$#2$}}%

\tcbset{
  takeaway/.style={ width=\hsize,left=0pt,right=0pt,top=0pt,bottom=0pt,colback=green!10!white,boxrule=1pt,colframe=black!30!green!50!white
  },
}
\newcounter{takeawaycounter}
\newcommand{\takeawaybox}[1]{
    \stepcounter{takeawaycounter}
    \begin{tcolorbox}[takeaway]
    \textbf{Takeaway~\arabic{takeawaycounter}:}#1
    \end{tcolorbox}
}

\tcbset{
  def/.style={ width=\hsize,left=1pt,right=1pt,top=1pt,bottom=1pt,colback=black!20!orange!4!white,boxrule=1pt,colframe=black!30!orange!20!white
  },
}

\title{\textcolor{green!60!black}{\faPagelines} Unveiling Environmental Impacts of Large Language Model Serving: \\ A Functional Unit View}

\author{
Yanran Wu, Inez Hua, Yi Ding \\
Purdue University \\
\texttt{\{wu2187, hua, yiding\}@purdue.edu}
}

\newcounter{question}[section]

\newcommand{\fuelq}[1]{
    \stepcounter{question} 
    \paragraph{Question \thequestion:} \textit{#1} 
}

\begin{document}
	
\maketitle

\begin{abstract}

Large language models (LLMs) offer powerful capabilities but come with significant environmental impact, particularly in carbon emissions. Existing studies benchmark carbon emissions but lack a standardized basis for comparison across different model configurations. To address this, we introduce the concept of \emph{functional unit} (FU) as a standardized basis and develop \SYSTEM{}, the first FU-based framework for evaluating LLM serving’s environmental impact. Through three case studies, we uncover key insights and trade-offs in reducing carbon emissions by optimizing model size, quantization strategy, and hardware choice, paving the way for more sustainable LLM serving. The code is available at \url{https://github.com/jojacola/FUEL}.


\end{abstract}

\section{Introduction} \label{sec:intro}

Large language models (LLMs) have been widely adopted in various industries due to their ability to perform complex language tasks~\cite{vu2023freshllms,shen2024hugginggpt,liu2024your}. However, LLM serving comes with significant environmental impact, particularly in terms of carbon emissions. For instance, processing a single prompt on ChatGPT produces over 4 grams of CO\textsubscript{2}eq~\cite{chatgptcarbon2023}, which is over 20× the carbon emissions generated by a web search query~\cite{whyyourinternet2020}.

Recent studies have benchmarked the carbon emissions of LLM serving by analyzing performance (e.g., throughput, latency) and energy consumption, then modeling carbon emissions under varying conditions such as request rate, and input/output length~\cite{nguyen2024towards,li2024towards,shi2024greenllm,li2024sprout}. However, these efforts have two limitations: \textbf{(1)} they focus on individual LLM rather than cross-model comparisons, and \textbf{(2)} they lack a standardized basis for carbon emission comparisons. These gaps limit the broader applicability and fairness of their analyses.

Building on principles from life cycle assessment in environmental sustainability~\cite{klopffer2014life}, we address these two limitations by introducing the concept of \emph{functional unit} (FU) as a standardized basis for comparing LLMs. In LLM serving, an FU represents a token generation defined by workload intensity, performance, and quality constraints. Using this, we develop \SYSTEM{}, a \underline{\textbf{F}}unctional \underline{\textbf{U}}nit-based \underline{\textbf{E}}valuation framework for evaluating the environment impact of \underline{\textbf{L}}LMs. To demonstrate its effectiveness and generalizability, we conduct three case studies exploring model size, quantization strategy, and hardware choice. Our key insights for building sustainable LLM serving systems include:  

\begin{itemize}
    \item \emph{Model size:} Larger models are greener in high output quality and low request rate, while smaller models excel as the request rate increases. 
    \item \emph{Quantization:} Quantization significantly lowers carbon emissions, especially for larger models.  
    \item \emph{Hardware:} Newer hardware offers better performance but is not always greener due to higher embodied carbon. Older hardware can lower carbon emissions while meeting quality and performance constraints under certain conditions.
\end{itemize}
The contributions of this paper are:
\begin{itemize}
    \item Introducing FU as a standardized basis for comparing different LLMs.  
    \item Designing and implementing \SYSTEM{}, the first FU-based framework for assessing the environmental impact of LLM serving.
    \item Conducting three case studies on model size, quantization, and hardware to draw insights in building sustainable LLM serving systems.
\end{itemize}

\section{Related Work}\label{sec:related} 

\begin{figure*}[!t]
    \centering
    \includegraphics[width=0.9\textwidth]{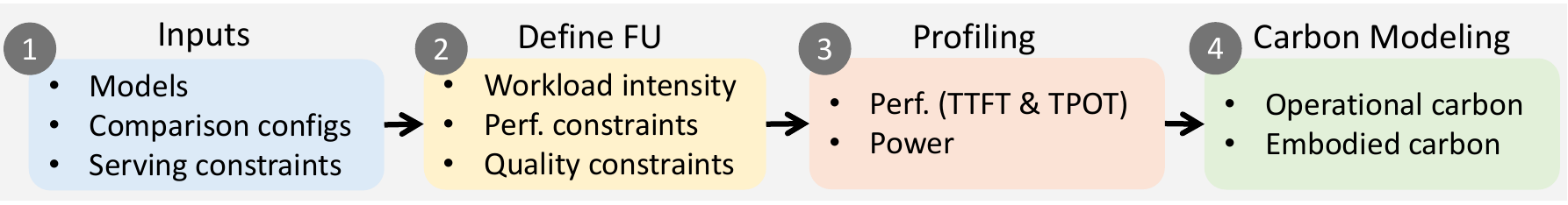}
    \caption{Overview of \SYSTEM{} framework.}
    \label{fig:framework}
\end{figure*}

\noindent \textbf{Environmental impact of LLM serving.} Researchers have recognized the environmental impact of LLM serving and explored it through modeling and profiling~\cite{ding2024sustainable}. Modeling efforts include LLMCarbon~\cite{faiz2024llmcarbon} and LLMCO2~\cite{fu2024llmco2}, which provide end-to-end carbon modeling frameworks, while LLMCampass~\cite{zhang2024llmcompass} focuses on hardware evaluation for LLM workloads. Profiling studies have run various LLM serving models across different hardware and QPS settings~\cite{nguyen2024towards,li2024arenahard,patel2024characterizing}, with GreenLLM~\cite{shi2024greenllm} and Sprout~\cite{li2024sprout} optimizing carbon emissions based on their profiling. However, none of these studies take a functional unit perspective as we do in this work.

\noindent \textbf{LLM serving optimization.} Prior work on LLM serving has primarily focused on optimizing performance and energy efficiency. Performance improvements can be categorized into model-level and system-level techniques. Model-side optimizations include quantization~\cite{lin2024awq, frantar2022gptq}, sparsification~\cite{frantar2023sparsegpt}, and speculative decoding~\cite{leviathan2023fast}. System-side approaches involve memory management~\cite{kwon2023efficient}, batching~\cite{agrawal2024taming, yu2022orca}, and kernel optimizations~\cite{dao2022flashattention}. Additionally, efforts to enhance energy efficiency include solutions like Splitwise~\cite{patel2024splitwise} and DynamoLLM~\cite{stojkovic2024dynamollm}. However, they have largely overlooked quality constraints when considering performance and energy efficiency.

\section{The Framework \SYSTEM{}} \label{sec:framework}

We present \SYSTEM{}, a \underline{\textbf{F}}unctional \underline{\textbf{U}}nit-based \underline{\textbf{E}}valuation framework for evaluating the environment impact of \underline{\textbf{L}}LMs. \SYSTEM{} enables a systematic and comprehensive analysis across various comparison configurations (e.g., model size, quantization, and hardware). Inspired by life cycle assessment in environmental sustainability~\cite{klopffer2014life}, the key insight is to establish a functional unit as a standardized basis for comparison. In LLM serving, a \emph{functional unit} (FU) represents a token characterized by its serving constraints during generation. In the \SYSTEM{} framework, we compare the environmental impact of tokens generated by different model configurations with the same performance and quality constraints.

\Cref{fig:framework} illustrates the four key steps of \SYSTEM{}. First, \SYSTEM{} identifies the inputs, including models, comparison configurations, and serving constraints. Next, it defines the FU based on these inputs. Then, experiments are conducted to profile performance and energy consumption. Finally, \SYSTEM{} quantifies the environmental impact --- focusing on carbon emissions in this work --- using the collected data. Next, we will introduce each step in detail.

\subsection{Step 1: Inputs}

The inputs to \SYSTEM{} include three key components:
\begin{itemize}
    \item \emph{Models:} The LLMs being compared, which can be different versions within the same model family or models from different families.  
    \item \emph{Comparison configurations:} The primary parameter that varies across comparisons. This paper focuses on three configurations: model size, quantization, and hardware.  
    \item \emph{Serving constraints:} The standardized basis for comparison, including workload intensity, performance constraint, and quality constraint. These constraints are critical in defining the FU.
\end{itemize}

\subsection{Step 2: Define Functional Unit}

In LLM serving, a \emph{functional unit} represents a token characterized by its workload intensity, performance, and quality constraints during generation. 

\noindent \textbf{Workload intensity.} \SYSTEM{} defines workload intensity as the request rate (QPS), measuring incoming user requests per second (req/s).

\noindent \textbf{Performance constraint.} \SYSTEM{} evaluates performance using two widely adopted metrics: Time-to-First-Token (TTFT) and Time-Per-Output-Token (TPOT). TTFT reflects how quickly the system responds to a new request by generating the first token, while TPOT quantifies the time per output token during decoding. Following prior work \citet{liu2024andes}, \SYSTEM{} sets a TTFT requirement of 1 second and a TPOT threshold of 200 ms, aligning with average human reading speed to ensure a smooth user experience.

\noindent \textbf{Quality constraint.} Quantitatively assessing output quality is challenging. While prior works (\citet{zhong2022unieval, yuan2021bartscore, jiang2023tigerscore}) have introduced various methods, they depend on either specific datasets or the need for reference answers. After evaluating multiple quality metrics, we adopt the reward model as a consistent quality evaluator for the model outputs across our experiments. The reward model is a common approach in reinforcement learning from human feedback (RLHF) training~\cite{ouyang2022training}. Specifically, we use the open-source pre-trained Skywork reward model~\cite{liu2024skywork}, which ranks highly on the RewardBench benchmark~\cite{lambert2024rewardbench}, reflecting its strong performance on diverse language generation tasks. Our experiments show that its reward scores align well with human preferences and effectively differentiate output quality across models. Using the reward model's score, we define \emph{Qscore} as a measure of output quality, where a higher Qscore reflects better quality and indicates that the output meets a certain quality threshold.

\noindent \textbf{An example of FU definition.} Based on these serving constraints, we define an example FU below:

\begin{tcolorbox}[def]
A token generated by an LLM at a request rate of 5 req/s, with a Qscore of 10, and performance constraints of 1s TTFT and 200ms TPOT.
\end{tcolorbox}

\subsection{Step 3: Profiling}

\SYSTEM{} profiles performance (TTFT and TPOT) and energy consumption by running LLMs under different configurations, based on the inputs given to \SYSTEM{} and the specified workload intensity. During profiling, Qscore is collected using an off-the-shelf reward model to evaluate output quality. For NVIDIA GPUs and Intel CPUs, power is measured every 200ms using NVIDIA (\verb|pynvml|) and Intel (\verb|psutil|) APIs for energy modeling, respectively.

\subsection{Step 4: Carbon Modeling}

Unlike prior work that profiles performance and energy without considering serving constraints, \SYSTEM{} defines and calculates \emph{carbon emission per FU} (CFU), measuring the emissions of FUs that meet certain serving constraints. Formally,
\[\Scale[0.99]{
\rm{CFU} = \frac{\rm{Total~carbon~emissions~for~all~tokens}}{N_{f}},}
 \]
\[ \Scale[0.9]{ N_{f} = \sum_{i=1}^N \mathbb{I}(\text{Q}_i \geq \alpha) \cdot \mathbb{I}(\text{TTFT}_i \leq \beta) \cdot \mathbb{I}(\text{TPOT}_i \leq \gamma) },\]


\noindent where $N$ is the total number of output tokens, \( N_{f} \) is the total number of tokens considered FUs, $\text{Q}$ is the Qscore, $\alpha$, $\beta$, and $\gamma$ are the constraints for Qscore, TTFT, and TPOT, respectively. Note that we consider a token to meet the Qscore requirement if its corresponding response does, as Qscore is defined at the response level. Next, we describe how to calculate carbon emissions.

\noindent \textbf{Carbon emission calculation.} Following prior work~\cite{nguyen2024towards,li2024towards,shi2024greenllm,ding2024sustainable},  total carbon emissions in LLM serving include operational carbon emission $C_{\rm op}$ and embodied carbon emissions $C_{\rm em}$. We now describe how to calculate each.
\begin{itemize}
    \item \emph{Operational carbon} is calculated as the product of the energy consumed, \( E_{\rm op} \), and the carbon intensity of the energy source (\(\texttt{CI}\)). Carbon intensity is the amount of $CO_{\rm 2eq}$ emitted per kilowatt-hour (\(kWh\)) of electricity used~\cite{maji2022carboncast,li2024uncertainty,yan2025ensembleci}. The operational carbon emission is thus given by:
\begin{align}\label{eq:carbon-op}
	C_{\rm op} = E_{\rm op} \cdot \texttt{CI}
\end{align}
\item \emph{Embodied carbon} of a hardware device is determined by factors such as processor chip area and memory capacity~\cite{gupta2022act,faiz2024llmcarbon}. The detail of modeling the total embodied carbon of a hardware device is in Appendix~\ref{sec:appendix_emb}. The embodied carbon emission of an LLM execution over time \( t \) is calculated by amortizing the hardware's total embodied carbon $C_{\rm em, total}$ over its lifetime ($\texttt{LT}$), typically 5 to 7 years~\cite{ostrouchov2020gpulife}. Thus, the embodied carbon for a time period \( t \) is given by:
\begin{align}\label{eq:carbon-eb}
	C_{\rm em} = \frac{t}{\texttt{LT}} \cdot C_{\rm em, total}
\end{align}
\item \emph{Total carbon} is thus given by:
\begin{align}\label{eq:carbon-eb}
	C_{\rm total} = E_{\rm op} \cdot \texttt{CI} + \frac{t}{\texttt{LT}} \cdot C_{\rm em, total}
\end{align}
\end{itemize}

\subsection{Summary and Implementation}

\SYSTEM{} provides a systematic framework for evaluating the environmental impact of LLM serving, using FU as a comparison basis. To demonstrate its effectiveness and generalizability, we will present three case studies exploring different comparison configurations: model size (\cref{sec:case1}), quantization (\cref{sec:case2}), and hardware (\cref{sec:case3}). For broadly applicable insights, we focus on two widely used model families, Qwen2.5~\cite{qwen2025qwen25technicalreport} and Llama2~\cite{touvron2023llama2openfoundation}, and conduct experiments using the open-source LLM serving platform vLLM~\cite{kwon2023efficient}. We use a carbon intensity of 518 gCO\textsubscript{2}eq, the 12-month average of our server's region, to calculate operational carbon emissions. All experiments were conducted in a single run with the LLM temperature set to 0 to minimize output randomness. We use the NewsQA \cite{trischler2016newsqa} summarization dataset for main results, as it tests language understanding without extra context. Results on other datasets are in the Appendix.

\section{Case Study: Model Size}\label{sec:case1} 

In this section, we use \SYSTEM{} to examine the environmental impact of model size on LLM serving.


\subsection{Evaluation Methodology}

\begin{figure}[!t]
    \centering
    \includegraphics[width=0.42\textwidth]{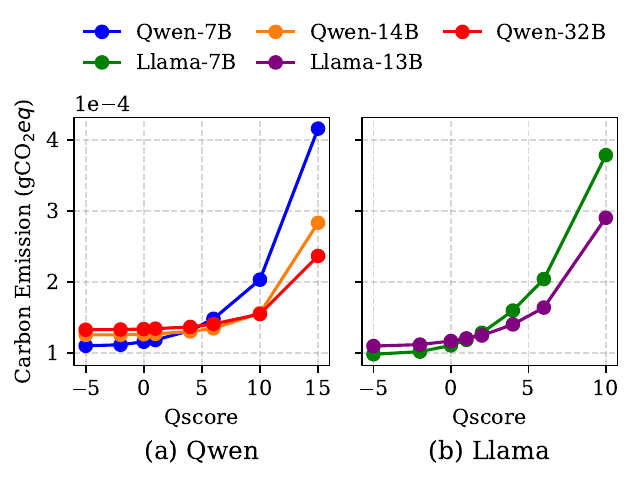}
    \vspace{-0.15in}
    \caption{Carbon emission per FU for different model sizes across Qscores at QPS=1 req/s.}
    \label{fig:size_carbon_per_token_quality}
\end{figure}
\begin{figure}[!t]
    \centering
    \includegraphics[width=0.49\textwidth]{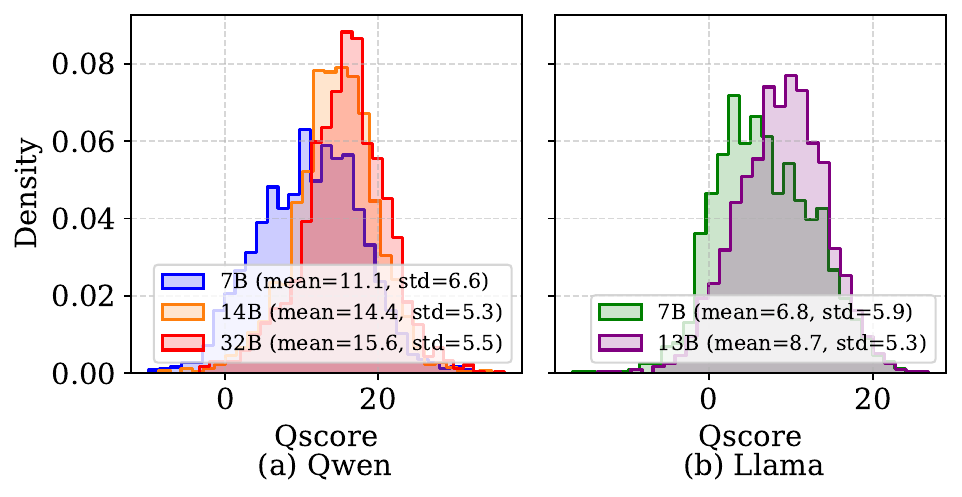}
    \vspace{-0.3in}
    \caption{Qscore distribution of outputs across different model sizes on the NewsQA dataset.}
    \label{fig:size_newsqa_score_dist}
\end{figure}

\paragraph{Setup.} We evaluate various model sizes from two LLM families---Qwen2.5 (7B, 14B, 32B) and Llama2 (7B, 13B)---on an NVIDIA H100 GPU paired with an Intel Xeon 8480+ CPU.

\paragraph{Benchmarking configurations.} To assess how model sizes affect the environmental impact---or how ``green" \textcolor{green!60!black}{\faPagelines} each model is in terms of carbon efficiency---we evaluate a range of FUs by adjusting serving constraints. QPS is from 1 to 20 req/s. The Qscore ranges are set to [-5, 15] for Qwen and [-5, 10] for Llama, based on the Qscore distribution of each model family (\Cref{fig:size_newsqa_score_dist} in \Cref{sec:appendix_model_size_newsqa}). These ranges ensure broad coverage while providing sufficient outputs across model sizes that meet quality requirements. TTFT is at 1s and TPOT is at 200ms to align with human reading speed.

\subsection{Evaluation Results}

\fuelq{Are smaller models always greener? }
We first investigate whether smaller models are always greener. \Cref{fig:size_carbon_per_token_quality} shows carbon emissions per FU across model sizes under different Qscore settings at QPS = 1 req/s. We choose a relatively low QPS to ensure all models generate enough tokens without violating performance constraints. The results indicate that the answer is \textbf{no}.  

For Qwen, at a low Qscore of -5, smaller models emit less carbon. However, as Qscore increases, carbon emissions increase for all model sizes, with smaller models increasing at a faster rate. When Qscore exceeds 5, the smallest 7B model becomes the highest emitter. At Qscore 15, the 32B model has the lowest emissions, while the 7B model emits over 1.8× more. A similar trend is seen in Llama, where larger models become greener as quality requirement rise. We confirm that larger models produce higher-quality outputs with higher Qscores in \Cref{fig:size_newsqa_score_dist}. This underscores the need to balance model size and output quality for lower carbon emissions.

These results demonstrate how FUEL provides a quantitative approach to assess the carbon efficiency of different models under consistent quality and performance requirements. In high-quality regimes, smaller models tend to emit more carbon per FU because fewer of their responses satisfy the criteria to be counted as functional units, leading to a higher average emission per FU.

\begin{figure}[!t]
    \centering
    \includegraphics[width=0.42\textwidth]{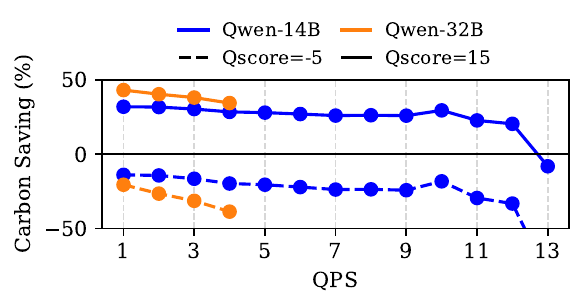}
    \vspace{-0.15in}
    \caption{Carbon savings of Qwen 14B and 32B compared to 7B with Qscore low (-5) and high (15). Data for Qwen 32B are missing at QPS > 4 req/s, as larger models cannot serve intensive workloads.}
    \label{fig:size_newsqa_cs_qwen}
\end{figure}

\begin{figure}[!t]
    \centering
    \includegraphics[width=0.42\textwidth]{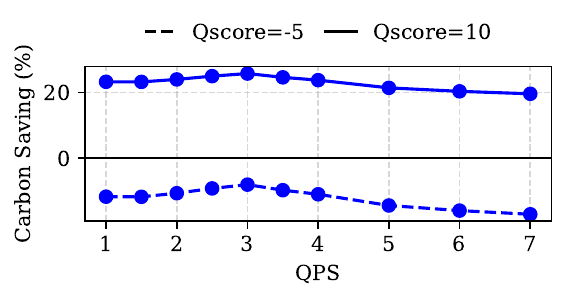}
    \vspace{-0.15in}
    \caption{Carbon savings of Llama 13B compared to 7B with Qscore low (-5) and high (10).}
    \label{fig:size_newsqa_cs_llama}
\end{figure}

\fuelq{When are larger models greener?}
To examine when larger models become greener, we set FUs with a broader QPS range and two quality requirements: low (Qscore = -5) and high (Qscore = 15 for Qwen, 10 for Llama). \Cref{fig:size_newsqa_cs_qwen} shows that for Qwen, larger models (14B and 32B) save more carbon compared to the 7B model under high Qscore, with the 32B saving over 40\%. However, under a low-quality requirement (Qscore = -5), larger models offer no advantage. A similar trend is seen for Llama, where the 13B model saves over 20\% carbon compared to the 7B model at high quality, as shown in \Cref{fig:size_newsqa_cs_llama}. Thus, \textbf{larger models become greener when output quality requirements are high}.

\begin{figure}[!t]
    \centering
    \includegraphics[width=0.42\textwidth]{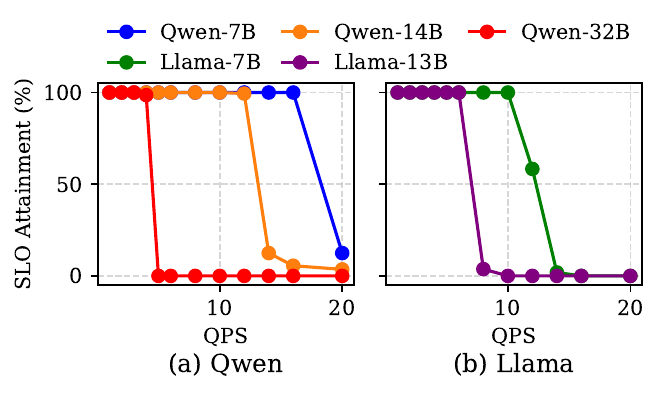}
    \vspace{-0.15in}
    \caption{SLO attainment of Qwen and Llama families across QPS range.}
    \label{fig:size_slo_attn}
\end{figure}

To explain the carbon savings shift with varying QPS, we analyze its impact on \emph{service level objective (SLO) attainment}, which refers to meeting TTFT and TPOT constraints. In \Cref{fig:size_slo_attn}, we observe that once QPS exceeds a certain threshold, SLO attainment drops, as the system becomes saturated. This explains why larger models can be greener at lower QPS: they meet performance constraints while producing higher-quality output.

\begin{figure}[!t]
    \centering
    \includegraphics[width=0.42\textwidth]{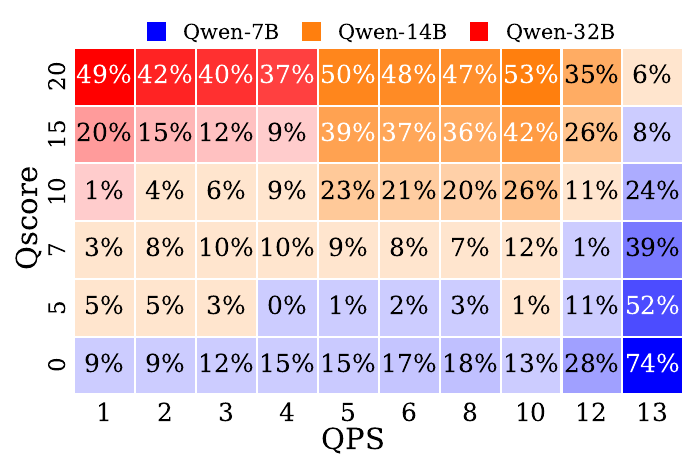}
    \vspace{-0.15in}
    \caption{Comparison of Qwen 7B, 14B, and 32B in FUEL. Tile colors indicate the model size with the lowest carbon per FU. Tile values are carbon savings (\%) of greenest model size compared to the second greenest.}
    \label{fig:size_heatmap}
\end{figure}

\fuelq{Does a universal greenest model size exist?}

\noindent The answer is \textbf{no}. \Cref{fig:size_heatmap} shows the relative carbon savings of Qwen 7B, 14B, and 32B across various QPS and Qscore values. No model size consistently has the lowest carbon emissions. At low QPS (1-4 req/s) with high Qscore, Qwen 32B can save up to 49\% in carbon emissions compared to the second greenest. However, as QPS increases, the 32B fails to meet the performance constraints, making the 14B the greenest. When the quality requirement is low (Qscore = 0), the 7B model is always the greenest, especially at high QPS.

\takeawaybox{
Larger models are greener under high-quality, low-QPS conditions. Smaller models become greener as QPS increases. No single model size is the greenest across all scenarios.
}

\section{Case Study: Quantization}\label{sec:case2} 

In this section, we explore how quantization affects the environmental impact of LLM serving. By reducing model weight and activation precision, quantization significantly decreases model size. For example, 4-bit quantization cuts model size by 4× compared to FP16. This reduction lowers memory usage and computational costs while maintaining accuracy. Using \SYSTEM{}, we investigate whether quantization, especially weight-only \cite{lin2024awq} and activation \cite{frantar2022gptq} quantization techniques, can improve carbon efficiency while maintaining output quality.

\subsection{Evaluation Methodology}

\paragraph{Setup.} We evaluate two widely used quantization methods: 4-bit AWQ \cite{lin2024awq} (weight-only) and W8A8 \cite{frantar2022gptq} (INT8 quantization for both weights and activations). We evaluate Qwen2.5 (7B, 14B, 32B) and Llama2 (7B, 13B) on an NVIDIA H100 GPU with an Intel Xeon 8480+ CPU. Qwen provides an official AWQ version, while Llama’s AWQ is from Hugging Face~\cite{huggingface_llama7_awq_2023, huggingface_llama13_awq_2023}. For W8A8, we quantize the models using LLM Compressor~\cite{vllm_llm_compressor_2023}, an open-source library designed for vLLM.

\paragraph{Benchmarking configurations.} Same as in \Cref{sec:case1}.

\subsection{Evaluation Results}

\fuelq{Is weight-only quantization always greener?}

\begin{figure}[!t]
    \centering
    \includegraphics[width=0.45\textwidth]{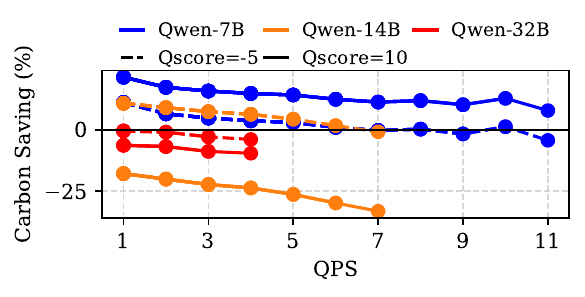}
    \vspace{-0.15in}
    \caption{Carbon savings of AWQ Qwen compared to the FP16 version with Qscore low (-5) and high (10). Data are missing at higher QPS for 14B and 32B, as larger models cannot serve intensive workloads.}
    \label{fig:quant_cs_qwen}
\end{figure}

\begin{figure}[!t]
    \centering
    \includegraphics[width=0.45\textwidth]{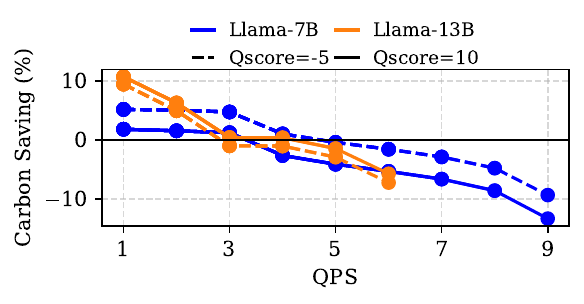}
    \vspace{-0.15in}
    \caption{Carbon savings of AWQ Llama compared to the FP16 version with Qscore low (-5) and high (10). Data are missing at higher QPS for 13B, as larger models cannot serve intensive workloads.}
    \label{fig:quant_cs_llama}
\end{figure}

\noindent The answer is \textbf{no}. \Cref{fig:quant_cs_qwen} shows the relative carbon emission savings per FU for AWQ compared to the FP16 version of Qwen under high (10) and low (-5) Qscores. Overall, AWQ’s carbon savings decline as QPS increases. For the 7B model, AWQ consistently reduces emissions, even under high Qscore. At QPS = 1 req/s and Qscore = 10, AWQ cuts emissions by over 20\% compared to FP16. This is because AWQ slightly increases the output quality of 7B (\Cref{tab:quant_qscore} in \Cref{sec:appendix_quant}), resulting in an increased number of FUs. On the other hand, the 14B model shows positive carbon savings at low Qscore (-5) but negative savings at high Qscore (10). The 32B model never achieves positive carbon savings, regardless of Qscore. We observe a similar trend for Llama in~\Cref{fig:quant_cs_llama}. As QPS increases, the carbon savings of AWQ over FP16 decline and can even become negative at high QPS. 

\begin{figure}[!t]
    \centering
    \includegraphics[width=0.45\textwidth]{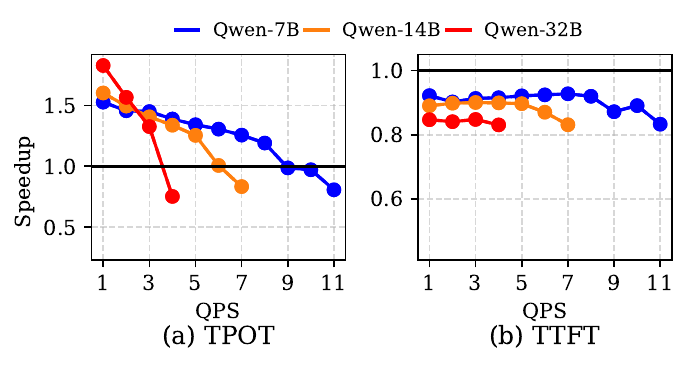}
    \vspace{-0.15in}
    \caption{Latency speedup of AWQ Qwen compared to the FP16 version.}
    \label{fig:quant_speedup_qwen}
\end{figure}

\begin{figure}[!t]
    \centering
    \includegraphics[width=0.45\textwidth]{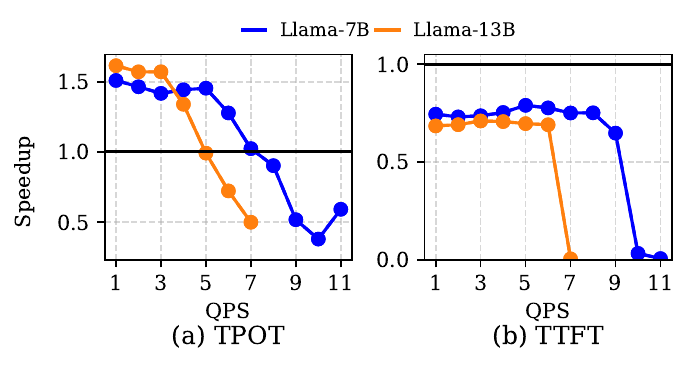}
    \vspace{-0.15in}
    \caption{Latency speedup of AWQ Llama compared to the FP16 version.}
    \label{fig:quant_speedup_llama}
\end{figure}

To understand why AWQ does not always outperform FP16 in carbon savings, we analyze its impact on TTFT and TPOT speedup. Figures \ref{fig:quant_speedup_qwen} and \ref{fig:quant_speedup_llama} show that TPOT sees some speedup at low QPS but slows down at high QPS, while TTFT is always slower than FP16. This is because quantization reduces weight size, but weights are dequantized back to 16-bit during inference, adding overhead. AWQ improves TPOT in memory-bound cases at low QPS by reducing memory transfer, but this advantage diminishes as QPS increases and computation grows. Since TTFT is compute-intensive, AWQ provides no speedup.

\takeawaybox{
Weight-only quantization reduces carbon emissions at low QPS but loses its advantage as QPS increases.
}



\fuelq{Is activation quantization always greener?}

\begin{figure}[!t]
    \centering
    \includegraphics[width=0.45\textwidth]{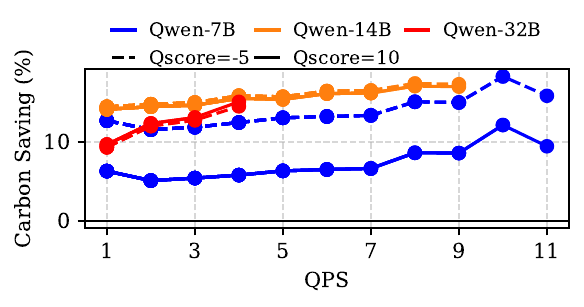}
    \vspace{-0.15in}
    \caption{Carbon savings of W8A8 Qwen compared to the FP16 version with Qscore low (-5) and high (10). Data are missing at higher QPS for 14B and 32B, as larger models cannot serve intensive workloads.}
    \label{fig:quant_cs_qwen_w8a8}
\end{figure}

\noindent Unlike weight-only quantization, activation quantization applies to both weights and activations. We compared the relative carbon savings of W8A8 compared to the FP16 version under different Qscores and QPS, and the results show that the answer is \textbf{yes}. As shown in~\Cref{fig:quant_cs_qwen_w8a8}, W8A8 consistently reduces carbon emissions for Qwen models, regardless of quality requirements. Despite some accuracy loss in the 7B model (\Cref{tab:quant_qscore} in \Cref{sec:appendix_quant}), it still achieves a 5\% carbon reduction at Qscore = 10. Unlike AWQ, W8A8 maintains stable savings even as QPS increases.

\begin{figure}[!t]
    \centering
    \includegraphics[width=0.45\textwidth]{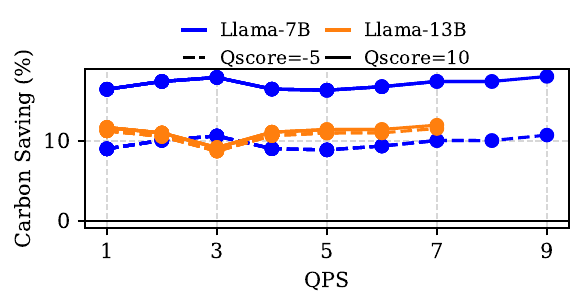}
    \caption{Carbon savings of W8A8 Llama compared to the FP16 version with Qscore low (-5) and high (10).}
    \vspace{-0.15in}
    \label{fig:quant_cs_llama_w8a8}
\end{figure}

We observe a similar trend for Llama in~\Cref{fig:quant_cs_llama_w8a8}. Notably, Llama 7B improved in output quality after quantization (\Cref{tab:quant_qscore} in \Cref{sec:appendix_quant}), saving over 15\% of carbon at Qscore = 10. This shows activation quantization can break the tradeoff between FP16 and AWQ, ensuring consistent carbon savings across different FUs.

\begin{figure}[!t]
    \centering
    \includegraphics[width=0.45\textwidth]{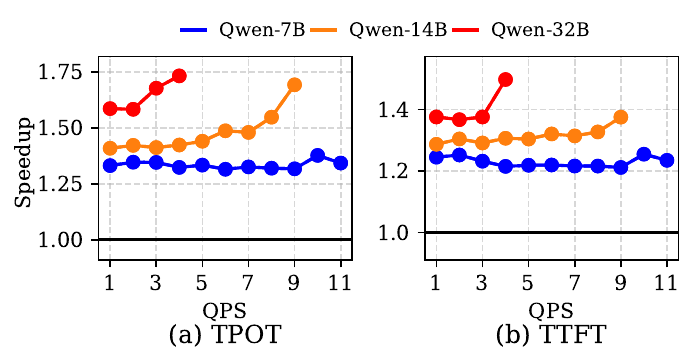}
    \vspace{-0.15in}
    \caption{Latency speedup of W8A8 Qwen compared to the FP16 version.}
    \label{fig:quant_speedup_qwen_w8a8}
\end{figure}

\begin{figure}[!t]
    \centering
    \includegraphics[width=0.45\textwidth]{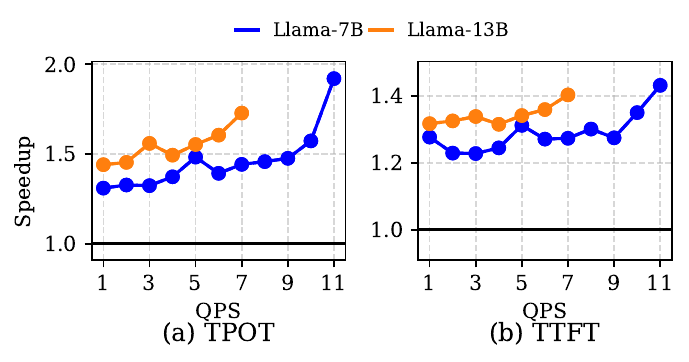}
    \vspace{-0.15in}
    \caption{Latency speedup of W8A8 Llama compared to the FP16 version.}
    \label{fig:quant_speedup_llama_w8a8}
\end{figure}

To understand why W8A8 always outperforms FP16 in carbon savings, we analyze its impact on TTFT and TPOT speedup. Figures \ref{fig:quant_speedup_qwen_w8a8} and \ref{fig:quant_speedup_llama_w8a8} show that W8A8 consistently speeds up TPOT and TTFT across all QPS ranges. This improvement comes from reducing both weight and activation precision, which decreases the amount of data movement and computation during inference. This makes W8A8 a more sustainable choice for LLM serving, as it strikes a balance between quality and performance.

\fuelq{Does a universal greenest quantization method exist?}

\begin{figure}[!t]
    \centering
    \includegraphics[width=0.48\textwidth]{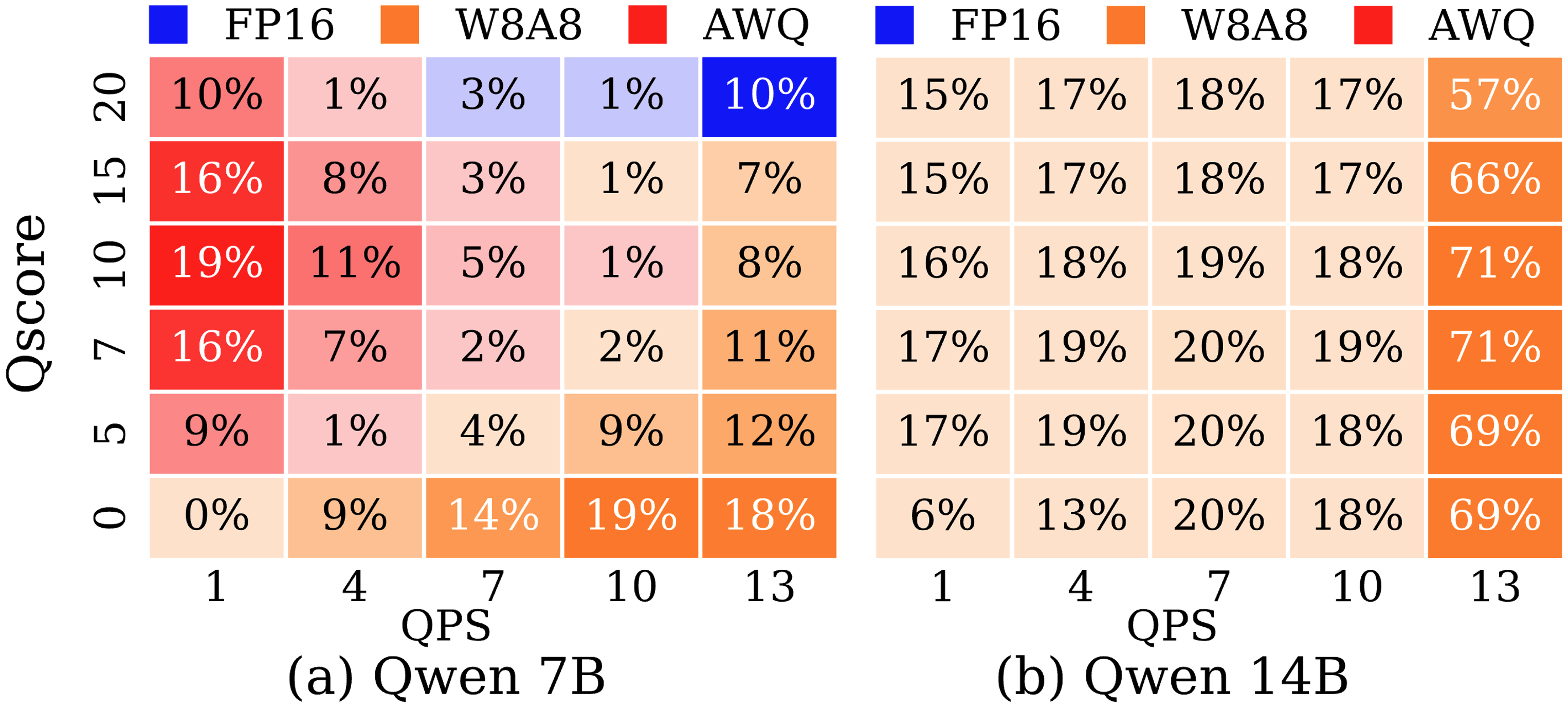}
    \vspace{-0.2in}
    \caption{Comparison of FP16, AWQ and W8A8 versions of Qwen 7B/14B in \SYSTEM{}. Tile colors indicate the model with the lowest carbon per FU. Tile values are carbon savings (\%) of greenest quantization versioncompared to the second greenest.}
    \label{fig:quant_heat_w8a8_qwen}
\end{figure}

\noindent The answer is \textbf{no}. \Cref{fig:quant_heat_w8a8_qwen} shows the relative carbon savings of FP16, AWQ, and W8A8 models across various QPS and Qscores for Qwen 7B and 14B. For Qwen 14B, W8A8 outperforms in all scenarios, with carbon savings increasing as QPS rises. However, for Qwen 7B, AWQ maintains slightly better quality at low QPS, while W8A8 lags behind at high QPS and high-quality requirements due to its slight accuracy loss (\Cref{tab:quant_qscore} in \Cref{sec:appendix_quant}).

\takeawaybox{Weight and activation quantization methods, like W8A8, hold significant potential for reducing carbon emissions in LLM serving, particularly for larger models. }



\section{Case Study: Hardware}\label{sec:case3} 

In this section, we examine how hardware platform affects the environmental impact of LLM serving. Using \SYSTEM{}, we investigate whether more advanced hardware can enhance carbon efficiency while maintaining output quality.

\subsection{Evaluation Methodology}

\paragraph{Setup.} We conduct experiments on two GPU servers with different hardware configurations, one older and one newer, as detailed in~\Cref{tab:hardware_specifications}. For fair comparisons, we use a single GPU per server for all experiments. We evaluate the Qwen2.5 (7B, 14B) and Llama2 (7B, 13B).

\paragraph{Benchmarking configurations.} Same as in \Cref{sec:case1}.

\begin{table}[!t]
    \centering
    \caption{Hardware platform specifications in this paper.}
    \footnotesize
    \begin{tabular}{l|p{2.5cm}|p{2.5cm}}
        \toprule
        \textbf{Specification} & \textbf{L40 server} & \textbf{H100 server} \\
        \midrule        
        \textbf{GPU} & 4 $\times$ L40 & 8$\times$ H100 \\
        \footnotesize{TDP} & \footnotesize{300W} & \footnotesize{700W} \\
        \footnotesize{Process size} & \footnotesize{5nm} & \footnotesize{5nm} \\
        \footnotesize{Die size} & \footnotesize{609 mm$^2$} & \footnotesize{814 mm$^2$} \\
        \footnotesize{GPU memory} & \footnotesize{40GB} & \footnotesize{80GB} \\
        \footnotesize{Release Year} & \footnotesize{2022} & \footnotesize{2023} \\
        \hline
        \textbf{CPU} & AMD EPYC 7443 & Intel Xeon 8480+ \\
        \footnotesize{TDP} & \footnotesize{200W} & \footnotesize{350W} \\
        \footnotesize{Process size} & \footnotesize{7nm} & \footnotesize{10nm} \\
        \footnotesize{Die size} & \footnotesize{4$\times$81 mm$^2$} & \footnotesize{4$\times$477 mm$^2$} \\
        \footnotesize{CPU memory} & 504GB & 1031GB \\
         \footnotesize{Release Year} & \footnotesize{2021} & \footnotesize{2023} \\
        \bottomrule
    \end{tabular}
    \label{tab:hardware_specifications}
\end{table}

\subsection{Evaluation Results}

\fuelq{How does different hardware contribute to total carbon emissions?}

\begin{figure}[!t]
    \centering
    \includegraphics[width=0.45\textwidth]{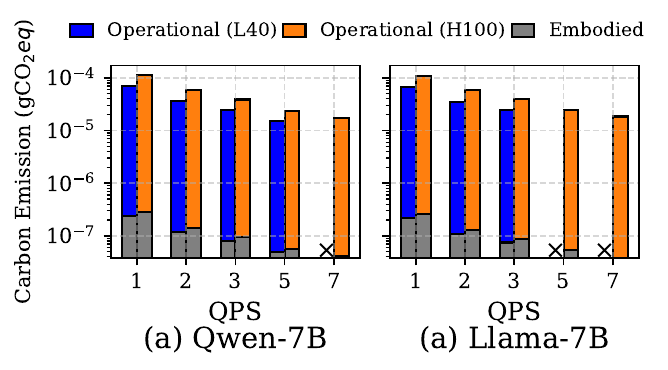}
    \vspace{-0.15in}
    \caption{Breakdown of Carbon emission per FU for Qwen and Llama 7B models on different hardware platforms, evaluated in FUEL with Qscore=0.}
    \label{fig:hw_carbon_breakdown}
\end{figure}

\noindent \Cref{fig:hw_carbon_breakdown} shows the breakdown of carbon emissions per FU for Qwen and Llama 7B models on different hardware platforms, separating operational and embodied carbon. It is worth noting that different hardware contributes to different embodied carbon per FU, due to differences in the total embodied carbon for each hardware. The L40 platform has lower total embodied carbon than the H100, with values of 26.6 and 29.92 kgCO\textsubscript{2}eq repectively. These differences are based on calculations using the ACT modeling tool~\cite{gupta2022act} and are due to hardware factors such as process and die size. The difference is even more pronounced between the AMD EPYC 7443 and Intel Xeon 8480+ CPUs, with the AMD CPU having 9.98 kgCO\textsubscript{2}eq, compared to the Intel's 42.81 kgCO\textsubscript{2}eq, over 4x higher.

Advanced hardware like the H100 offers better performance but higher embodied carbon. Extending hardware lifetime can yield more carbon savings, especially considering the large difference in embodied carbon between older and newer devices.

\fuelq{Is LLM serving on advanced hardware greener?}

\begin{figure}[!t]
    \centering
    \includegraphics[width=0.42\textwidth]{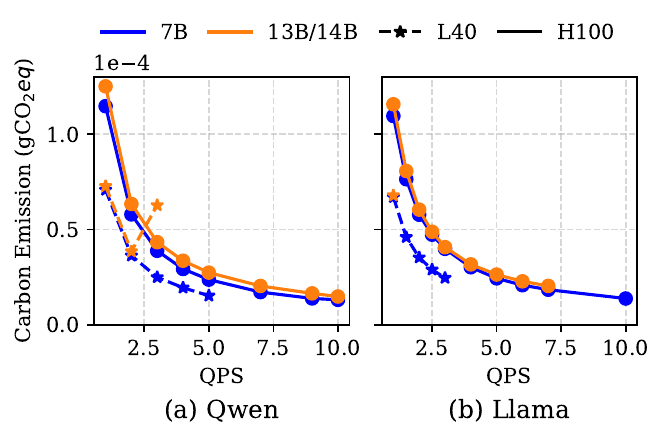}
    \vspace{-0.15in}
    \caption{Carbon emission per FU of Qwen and Llama model families on different hardware platforms, evaluated in FUEL with Qscore=0.}
    \label{fig:hw_carbon_models}
\end{figure}

\noindent \Cref{fig:hw_carbon_models} shows the carbon emissions per FU for the Qwen and Llama model families on two hardware platforms. At low QPS, the L40 server consistently has lower carbon emissions than the H100. This means that the answer is \textbf{no}: advanced hardware is not necessarily greener.

\begin{figure}[!t]
    \centering
    \includegraphics[width=0.42\textwidth]{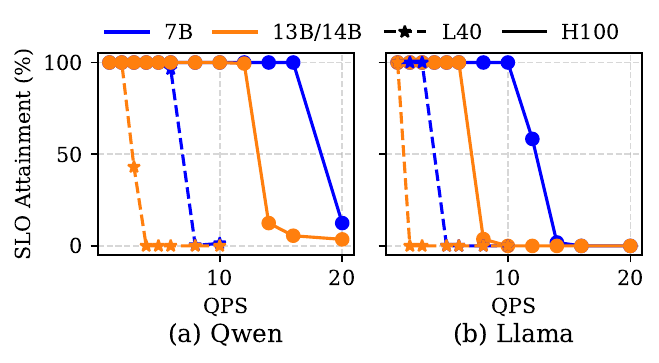}
    \vspace{-0.15in}
    \caption{SLO attainment of Qwen and Llama model families on different hardware platforms.}
    \label{fig:hw_slo_attn}
\end{figure}

The main advantage of advanced hardware such as the H100 lies in its higher computational throughput, memory capacity, and improved model support. These capabilities enable it to produce higher-quality outputs while satisfying latency and throughput constraints, as shown in~\Cref{fig:hw_slo_attn}. However, this comes at the cost of significantly higher power consumption and embodied carbon. As a result, advanced hardware does not always lead to lower emissions per FU, especially when older hardware is sufficient to meet the same quality and performance requirements.

\begin{figure}[!t]
    \centering
    \includegraphics[width=0.48\textwidth]{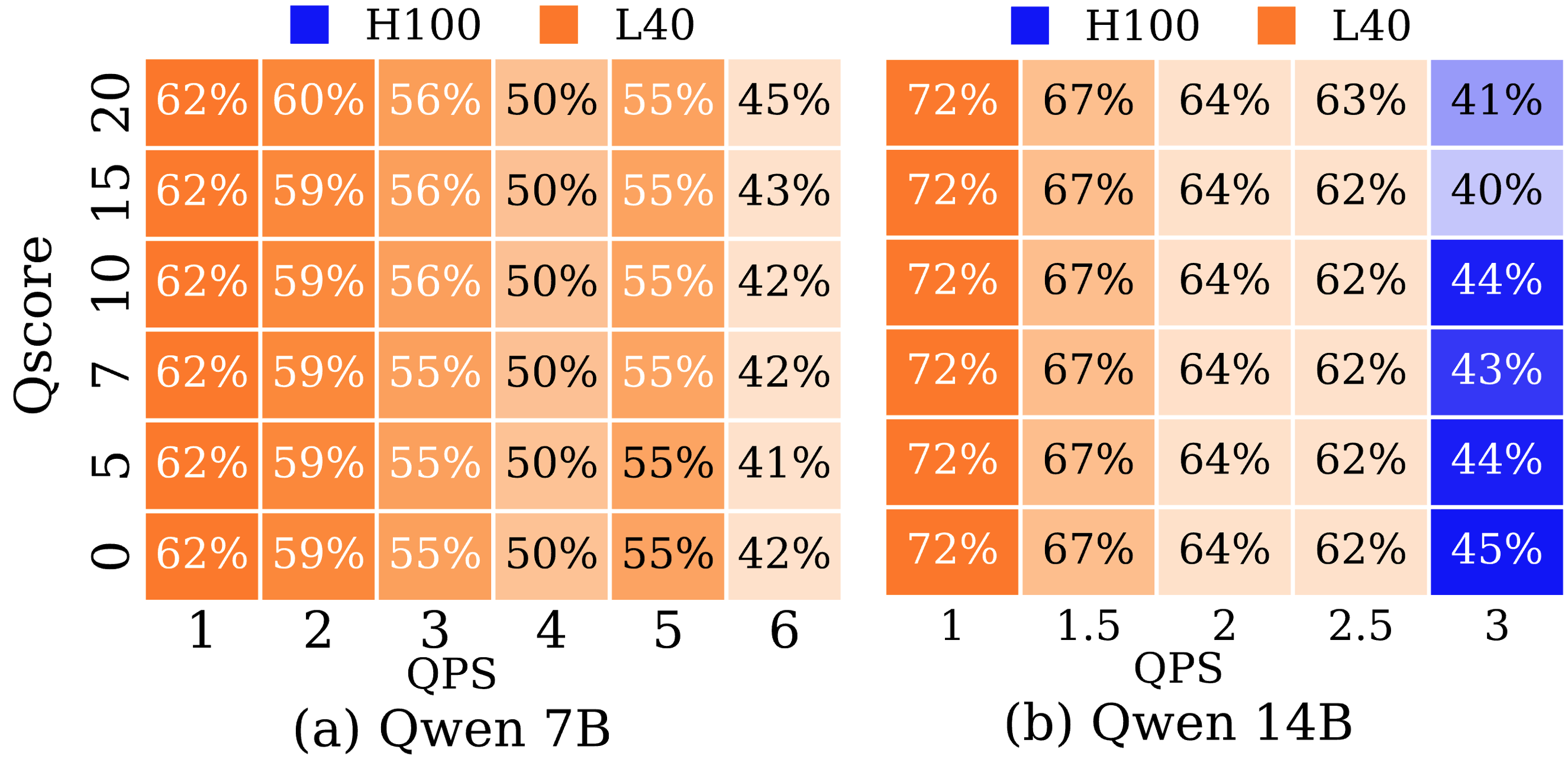}
    \vspace{-0.2in}
    \caption{Comparison of Qwen 7B and 14B on different hardware platforms in FUEL. Tile colors indicate the hardware with the lowest carbon emission per FU. Tile values are carbon savings (\%) of the greenest hardware compared to the second greenest.}
    \label{fig:hw_heatmap_qwen_sep}
\end{figure}

\fuelq{How to choose greener hardware?}

\noindent To answer this question, we run experiments across different FUs with varying QPS and Qscores. \Cref{fig:hw_heatmap_qwen_sep} shows the relative carbon savings of L40 and H100 servers for Qwen 7B and 14B. Hardware carbon efficiency depends mainly on model size and QPS, with a minor influence from Qscore. Newer hardware is more carbon efficient at high QPS, while older hardware is better at low QPS. These findings underscore the sustainability benefits of reusing older hardware to cut carbon emissions while maintaining performance and quality.

\takeawaybox{
Advanced hardware offers higher performance but is not always greener due to higher embodied carbon and power consumption. When quality and performance constraints can be met, older hardware can achieve lower carbon emissions per functional unit.
}

\section{Conclusion} \label{sec:conclusion}

We introduce \SYSTEM{}, the first evaluation framework for unveiling LLM serving's environmental impact by leveraging functional units as the basis for comparison. We explore how model size, quantization, and hardware affect carbon emissions. Our findings highlight opportunities for greener LLM deployment, paving the way for sustainable AI systems.



\section*{Acknowledgments}

This work is supported by NSF CCF-2413870.

\section*{Limitations}

We discuss the limitations of this work as follows.

\paragraph{Model families.} Our case studies examine two widely used open-source LLM families, Qwen2.5 and Llama2, which we believe are representative of general LLM serving behaviors. However, we have not yet explored other model families, such as Mistral, or task-specific models like multimodal, vision-language, and code-focused LLMs. We leave these investigations for future work.

\paragraph{Hardware.} All our experiments are conducted on a single GPU to ensure fair comparisons, limiting us to models up to 32B. We have yet to explore the performance and power dynamics in a multi-GPU distributed environment, which would allow us to run larger models like Llama 70B. This setup introduces additional overhead, particularly from communication, making the results even more insightful. We leave this exploration for future work.

\paragraph{Quality metrics.} Quantitatively evaluating LLM output quality remains a challenging and open research question. We experimented with various metrics before selecting the reward model, a common approach in reinforcement learning from human feedback. While we believe our key findings remain robust regardless of the specific quality metric used, access to more advanced evaluation methods in the future could further enhance the accuracy and rigor of our work.

\section*{Ethical Statement}

This work contributes to the development of carbon-efficient LLM serving systems. We are committed to conducting our work in a responsible manner, adhering to ethical guidelines and best practices. 

We recognize the potential environmental impact of the widespread use of LLMs, including energy consumption, electronic waste, and the environmental impact of hardware manufacturing. Therefore, we emphasize the importance of optimizing LLMs for lower energy and carbon emissions, not only in terms of performance but also through hardware reuse and longevity, as part of a more sustainable approach to AI infrastructure.

We are transparent in our research methodologies. As we explore new avenues for improving LLM efficiency, we remain mindful of the broader social, economic, and environmental implications of deploying large-scale AI systems and aim to promote solutions that benefit both the technology and society at large.

We also recognize the importance of fairness and inclusivity, ensuring that our research does not disproportionately harm any community or group and aligns with the goal of creating AI systems that are accessible and beneficial to all.


\bibliography{refs}

\appendix

\section{Appendix Overview}

We summarize the appendix as follows:

\begin{itemize}
    \item Section~\ref{sec:appendix_emb} provides a detailed description of embodied carbon modeling.
    \item Section~\ref{sec:appendix_model_size} presents additional results for the model size case study, including experiments on NewsQA and two additional datasets (Arena Hard and HumanEval).
    \item Section~\ref{sec:appendix_quant} provides supplementary results for the quantization case study on the same datasets.
    \item Section~\ref{sec:appendix_hardware} offers more detailed comparisons of model selections in the hardware case study.
\end{itemize}


\section{Embodied Carbon Modeling}\label{sec:appendix_emb}
We utilize the ACT~\cite{gupta2022act} embodied carbon modeling tool. The embodied carbon footprint can be divided into manufacturing and packaging carbon emissions. Manufacturing carbon arises from producing electronic components like transistors and resistors from raw materials, while packaging carbon is associated with assembling these components into chips and circuit boards:
\begin{equation}
    C_{\rm em} = C_{\rm manufacturing} + C_{\rm packaging}
\end{equation}

The manufacturing embodied footprint $C_m$ of processors and SoCs like CPUs and GPUs depends on several factors: die area ($A_{\text{die}}$), carbon intensity of the energy consumed by the fab ($\texttt{CI}_{\rm fab}$), energy consumed per unit area manufactured (EPA), the GHG emissions from gases and chemicals per unit area (GPA), the footprint of procuring raw materials per unit area (MPA), and fabrication yield ($\text{Yield}$, set to 0.875 as in \citet{gupta2022act}). The information is sourced from product data sheets and sustainability reports. The manufacturing embodied carbon of a processor can be calculated as:

\begin{equation}
    C_{\rm m} = \frac{(\texttt{CI}_{\rm fab} \times \text{EPA} + \text{GPA} + \text{MPA})\times A_{die}}{\text{Yield}}
\end{equation}

The packaging carbon emission $C_{\rm p}$ is calculated by the number of integrated circuits ($N_{\rm IC}$) with a packaging footprint. Following ACT,  we use an average packaging overhead of 150 gCO\textsubscript{2}eq per IC.
\begin{equation}
    C_{\rm p} = N_{\rm IC} \times 150
\end{equation}

In cloud environments or HPC clusters, it is often challenging to obtain details of DRAM specifications. Previous studies~\cite{li2023toward, mem2023} generally assume that the embodied carbon of DRAM is proportional to its capacity. Following prior work, we adopt a fixed rate of 65 gCO\textsubscript{2}eq/GB to estimate the embodied carbon of DRAM.

\section{Additional Results for Model Size Case Study}\label{sec:appendix_model_size}

\subsection{Results on NewsQA Summarization}
\label{sec:appendix_model_size_newsqa}
\begin{figure}[!t]
    \centering
    \includegraphics[width=0.48\textwidth]{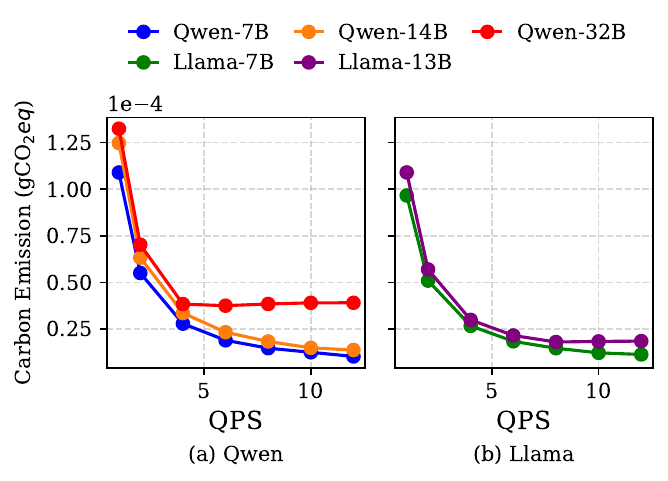}
    \caption{Naive carbon emission per token for different model sizes across QPS range on NewsQA dataset.}
    \label{fig:size_carbon_per_token}
\end{figure}

Figure~\ref{fig:size_carbon_per_token} illustrates the naive carbon emission per token for various model sizes across a range of QPS. This figure represents carbon per token without the use of \SYSTEM{}. Without considering server constraints, smaller models consistently exhibit lower carbon emissions per token, which does not reflect real-world serving requirements where larger models may be preferred for higher quality outputs.

\begin{figure}[!t]
    \centering
    \includegraphics[width=0.48\textwidth]{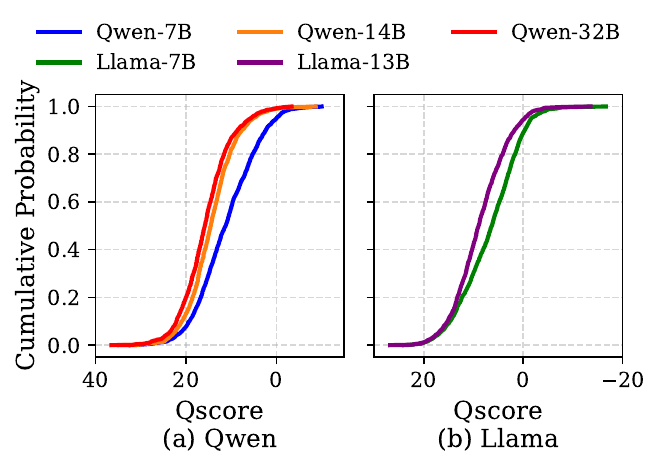}
    \caption{Cumulative percentage of Qscore $\geq$ threshold on NewsQA dataset.}
    \label{fig:size_newsqa_score_cdf}
\end{figure}

\Cref{fig:size_newsqa_score_cdf} shows the cumulative percentage of quality scores $\geq$ a given threshold for different models on the NewsQA summarization task. This figure highlights significant differences between models, particularly between Llama 7B and 13B, and between Qwen 7B and 32B. This discrepancy demonstrates why smaller models may not be as advantageous when higher quality is required, as larger models provide better outputs.

\subsection{Results on Arena Hard}
The Arena Hard dataset~\cite{li2024arenahard} is a challenging benchmark designed to evaluate the instruction following capabilities of LLMs, which is derived from real user interactions on Chatbot Arena.

\begin{figure}[!t]
    \centering
    \includegraphics[width=0.49\textwidth]{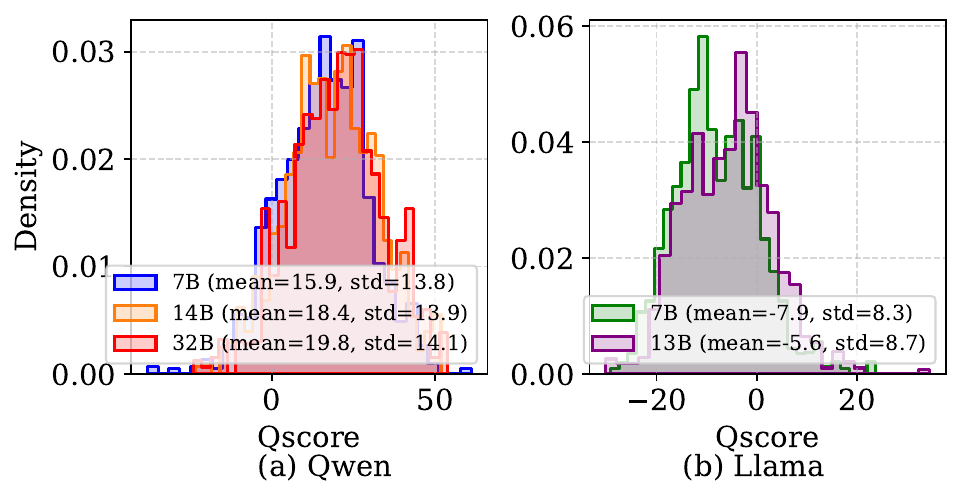}
    \caption{Qscore distribution of outputs across different model sizes on the Arena Hard dataset.}
    \label{fig:arena_size_score_dist}
\end{figure}

\begin{figure}[!t]
    \centering
    \includegraphics[width=0.49\textwidth]{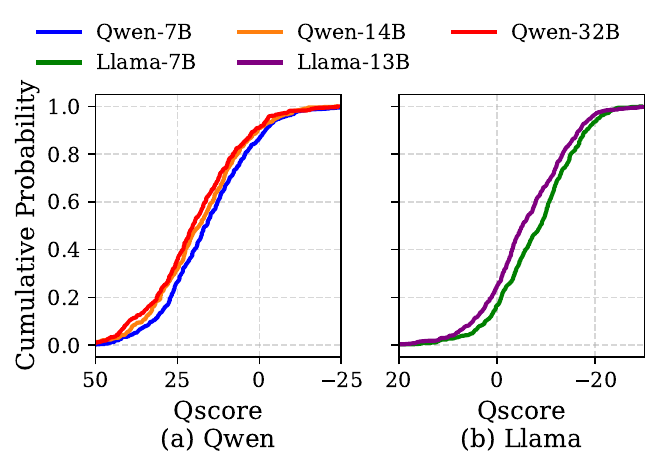}
    \caption{Cumulative percentage of Qscore $\geq$ threshold for different model sizes on Arena Hard dataset.}
    \label{fig:arena_size_score_dist_cumu}
\end{figure}

Figure~\ref{fig:arena_size_score_dist} shows the Qscore distribution for different model sizes on the Arena Hard dataset. As shown, larger models tend to achieve higher quality scores. However, compared to the quality distribution on the NewsQA summarization task, while larger models still perform better, the differences between model sizes on Arena Hard are less pronounced than on the NewsQA. However, the gap between Llama 7B and 13B remains significant. Figure~\ref{fig:arena_size_score_dist_cumu} also confirms this trend by showing the cumulative percentage of quality scores $\geq$ a given threshold for various model sizes on Arena Hard. 

\begin{figure}[!t]
    \centering
    \includegraphics[width=0.47\textwidth]{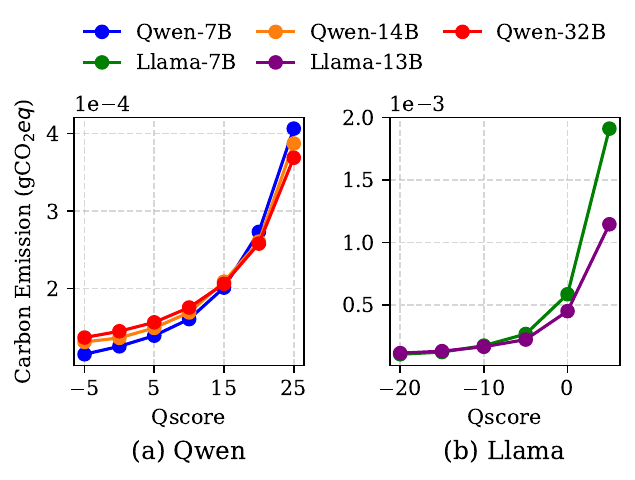}
    \caption{Carbon emission per FU for different model sizes on Arena Hard across Qscores at QPS=1 req/s.}
    \label{fig:arena_size_carbon_per_token_quality}
\end{figure}

\Cref{fig:arena_size_carbon_per_token_quality} shows carbon emissions per FU across model sizes under different Qscore settings at QPS = 1 req/s. For the Qwen model family, since the Qscore distribution gap has narrowed, we only observe the 32B model producing less carbon than the 7B model when the quality requirement becomes very high (Qscore > 15). On the other hand, due to the significant quality distribution gap between the Llama models, a slight increase in the quality requirement makes the Llama 13B model greener than the 7B model. Moreover, when the quality requirements become stricter, the carbon emission gap between Llama 13B and 7B becomes larger.

\begin{figure}[!t]
    \centering
    \includegraphics[width=0.48\textwidth]{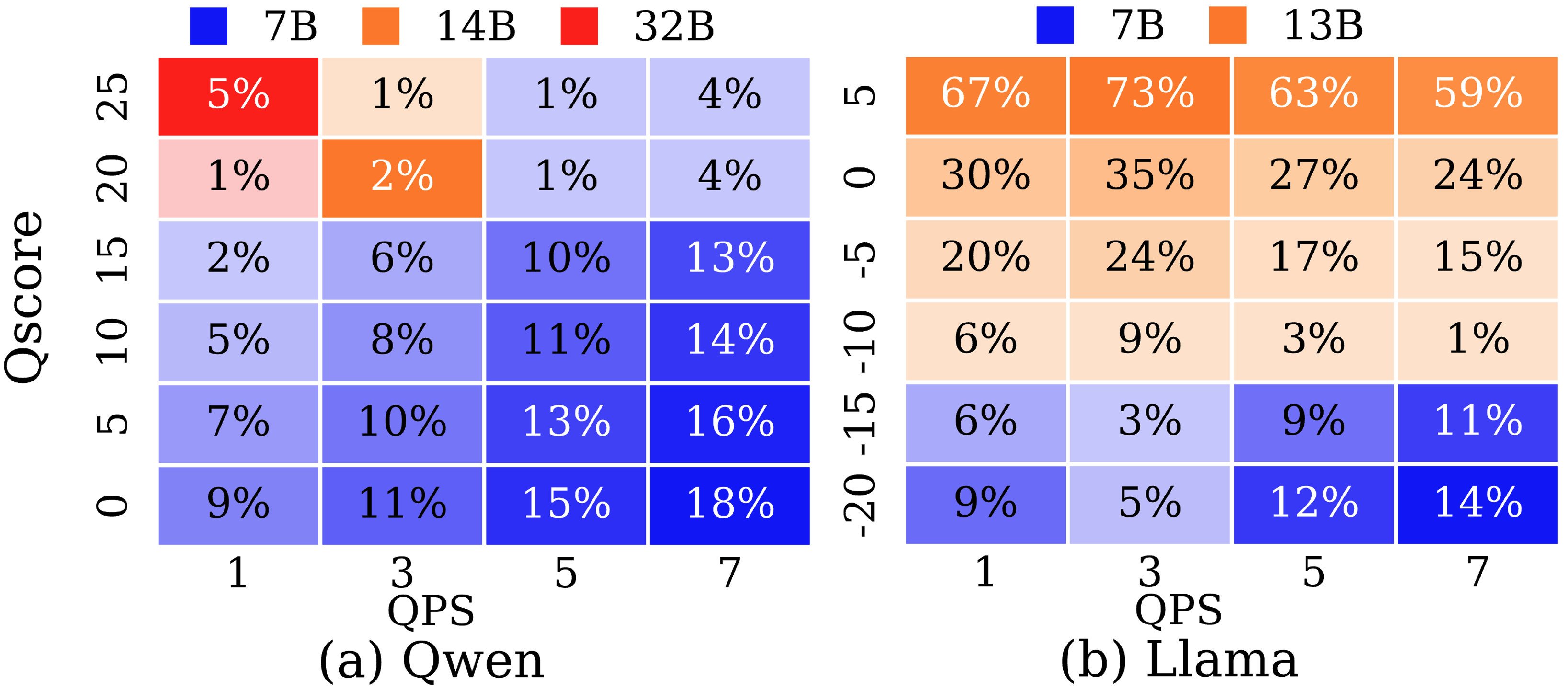}
    \caption{Comparison of different model sizes on Arena Hard in \SYSTEM{}. Tile colors indicate the model with the lowest carbon per FU. Tile values are carbon savings (\%) of the greenest size compared to the second greenest.}
    \label{fig:size_arena_heatmap}
\end{figure}

The result aligns well with our findings on the NewsQA dataset: if the quality requirement is high, larger models become a greener choice, especially when there are large differences in quality distribution across models of different sizes. \Cref{fig:size_arena_heatmap} shows the optimal model size choice across various Qscore and QPS conditions for the Qwen and Llama families. Due to the close quality distribution within the Qwen family, the advantage of larger models is constrained to the top-left corner (high Qscore, low QPS). In contrast, in the bottom right corner, as the quality requirement decreases and QPS increases, the 7B model becomes the greenest one.

\subsection{Results on HumanEval}
The HumanEval dataset~\cite{chen2021humaneval} is a benchmark designed to evaluate the code generation ability of LLMs. It consists of Python coding problems and requires LLMs to implement the specific functions. 

\begin{figure}[!t]
    \centering
    \includegraphics[width=0.49\textwidth]{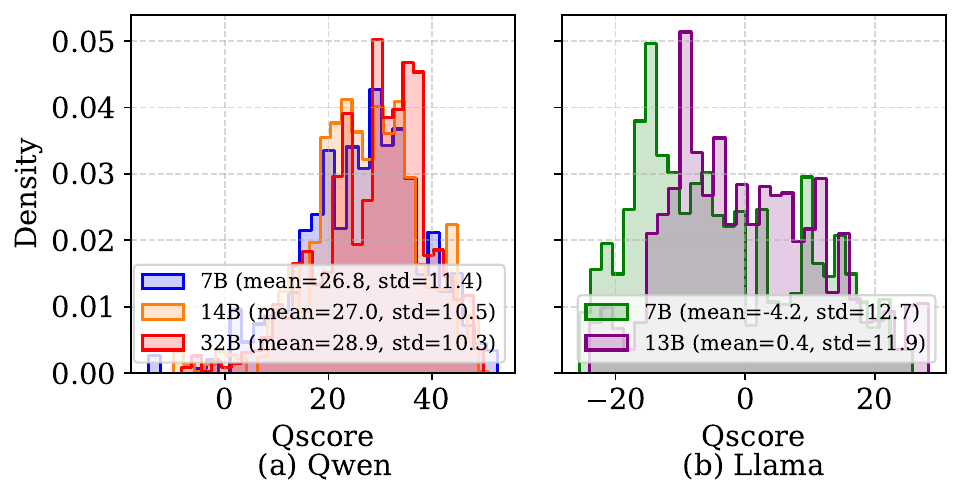}
    \caption{Qscore distribution for different model sizes on HumanEval dataset.}
    \label{fig:humaneval_size_score_dist}
\end{figure}

\Cref{fig:humaneval_size_score_dist} shows the Qscore distribution for different model sizes on the HumanEval dataset. Qscore distribution for the Qwen models is much closer on this dataset. This is consistent with their technical report~\cite{qwen2025qwen25technicalreport}, which also highlights similar performance across models in the HumanEval evaluation. However, for the Llama models, the gap between the 7B and 13B model remains large, as expected.

\begin{figure}[!t]
    \centering
    \includegraphics[width=0.49\textwidth]{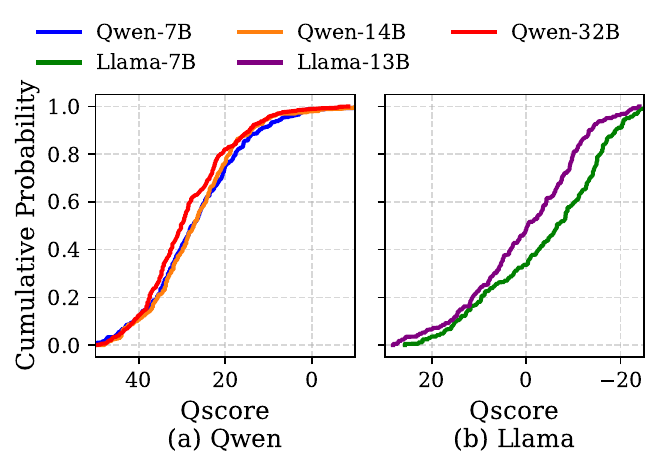}
    \caption{Cumulative percentage of Qscore $\geq$ threshold for different model sizes on HumanEval dataset.}
    \label{fig:humaneval_size_score_dist_cumu}
\end{figure}

\begin{figure}[!t]
    \centering
    \includegraphics[width=0.47\textwidth]{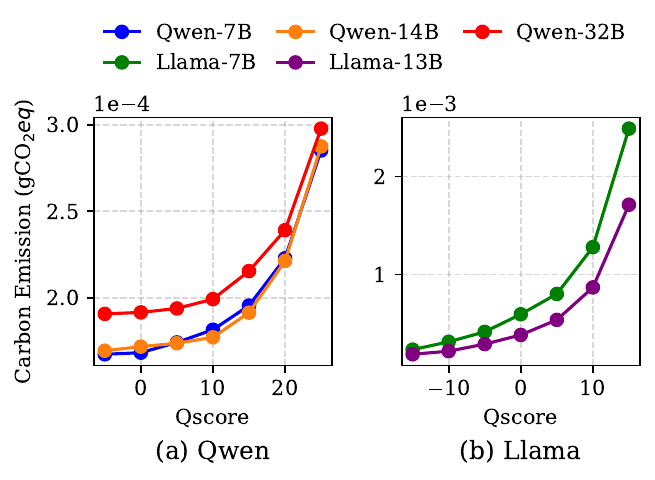}
    \caption{Carbon emission per FU for different model sizes on HumanEval across Qscores at QPS=1 req/s.}
    \label{fig:humaneval_size_carbon_per_token_quality}
\end{figure}

\Cref{fig:humaneval_size_carbon_per_token_quality} shows carbon emissions per FU across model sizes on HumanEval dataset under different Qscore settings at QPS = 1 req/s. For the Qwen family, since the quality difference between the three model sizes on this task is not significant, increasing the Qscore does not lead to larger models demonstrating carbon emission saving over the 7B model. The carbon emissions per FU remain similar across model sizes even with higher Qscore requirements.
In contrast, for the Llama model family, due to the large quality gap between the 7B and 13B models, we observe that even at very low quality requirements (e.g., Qscore = -15), the 13B model exhibits lower carbon emissions than the 7B model.

\begin{figure}[!t]
    \centering
    \includegraphics[width=0.48\textwidth]{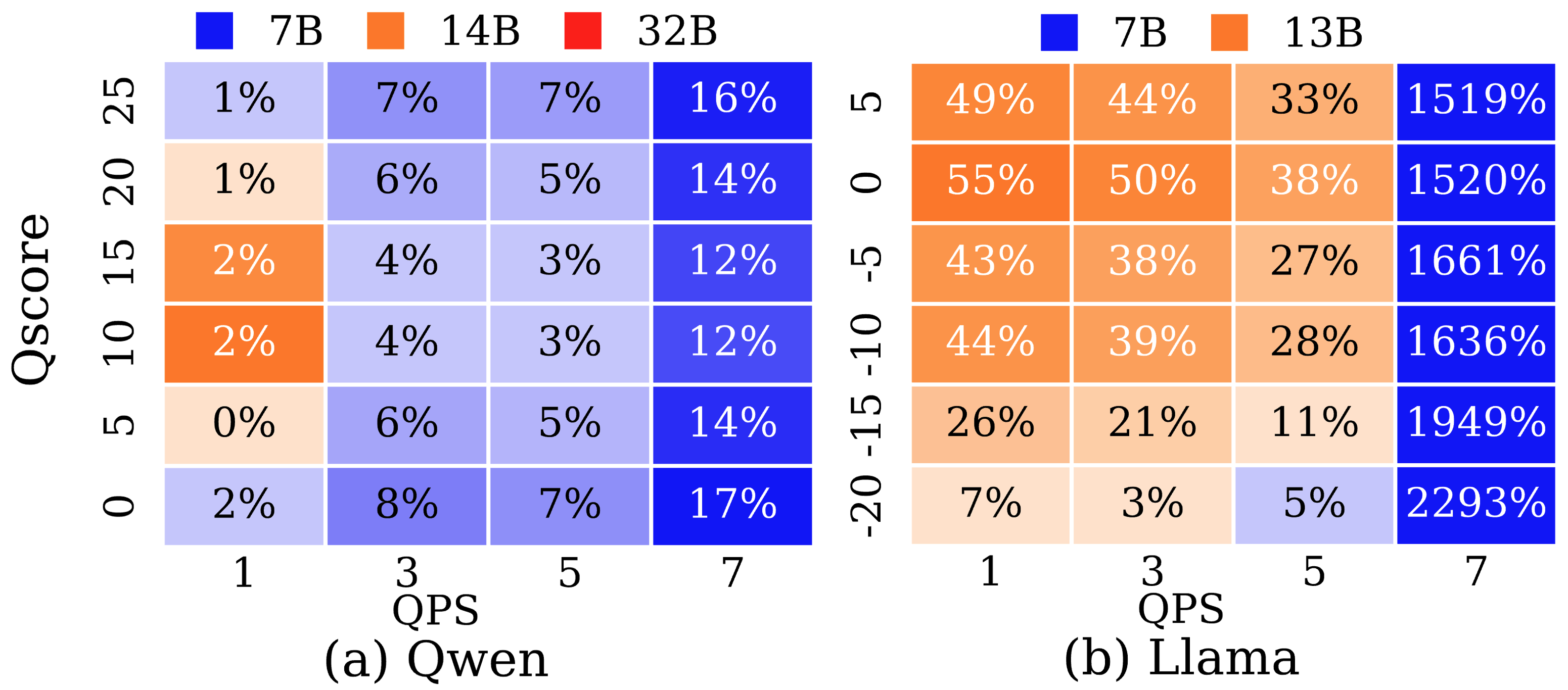}
    \caption{Comparison of different model sizes on HumanEval in \SYSTEM{}. Tile colors indicate the model with the lowest carbon per FU. Tile values are carbon savings (\%) of the greenest size compared to the second greenest.}
    \label{fig:humaneval_size_heatmap}
\end{figure}

If we extend the Qscore requirement and QPS into two dimensions, as demonstrated in \Cref{fig:humaneval_size_heatmap}, we observe that on HumanEval, Qwen 14B only shows an incremental carbon saving of 1-2\% at QPS = 1 req/s, while in most other cases, Qwen 7B remains the greenest model. This is because the output quality of Qwen 7B is very close to that of Qwen 14B and 32B. For the Llama model family, the results align with the previous observations: at lower QPS and higher quality requirements, the 13B model becomes the greenest option, as it can produce higher-quality responses compared to the 7B model.

This experiment on the HumanEval dataset further highlights our previous conclusion that selecting the greenest model size requires a comprehensive consideration of both model output quality and workload intensity.

\section{Additional Results for Quantization Case Study}
\label{sec:appendix_quant}

\begin{table*}[!t]
\centering
\caption{Evaluation on different benchmarks for Qwen and Llama families with their quantized versions.}
\label{tab:benchmark_quant}
\begin{adjustbox}{width=\textwidth}
\normalsize
\begin{tabular}{c|c|c|c|c|c|c|c}
\toprule
\textbf{Model} & \textbf{Method}  & \textbf{ARC-c} & \textbf{GSM8k } & \textbf{HellaSwag} & \textbf{MMLU } & \textbf{TruthfulQA} & \textbf{Winogrande}  \\
\midrule
\multirow{3}{*}{\textbf{Qwen-7B}} & FP16  & 63.57 & 81.96 & 62.24 & 74.23 & 49.82 & 73.64 \\
 & AWQ  & 62.03 {\color{red} (-1.54)} & 79.61 {\color{red} (-2.35)} & 61.52 {\color{red} (-0.72)} & 73.33 {\color{red} (-0.9)} & 50.43 {\color{green} (+0.61)} & 74.11 {\color{green} (+0.47)} \\
 & W8A8  & 63.65 {\color{green} (+0.08)} & 82.11 {\color{green} (+0.15)} & 62.15 {\color{red} (-0.09)} & 74.18 {\color{red} (-0.05)} & 49.45 {\color{red} (-0.37)} & 74.35 {\color{red} (-0.71)} \\
\hline
\multirow{3}{*}{\textbf{Qwen-14B}} & FP16  & 69.54  & 79.23 & 65.73 & 79.87 & 52.26 & 80.66 \\
 & AWQ   & 68.00 {\color{red} (-1.54)}& 80.89{\color{green} (+1.66)} & 64.78 {\color{red} (-0.95)} & 78.88 {\color{red} (-0.99)} & 48.84 {\color{red} (-3.42)} & 79.48 {\color{red} (-1.18)} \\
 & W8A8  & 69.71 {\color{green} (+0.71)} & 79.83 {\color{red} (-0.6)} & 65.74 {\color{green} (+0.01)} & 79.93 {\color{green} (+0.06)} & 51.04 {\color{red} (-1.22)} & 81.14 {\color{green} (+0.48)} \\
\hline
\multirow{3}{*}{\textbf{Qwen-32B}} & FP16  & 71.42  & 75.89 & 67.11 & 83.28 & 51.16 & 80.03 \\
 & AWQ   & 69.88 {\color{red} (-1.54)}& 76.72{\color{green} (+0.83)} & 66.47 {\color{red} (-0.64)} & 82.40 {\color{red} (-0.88)} & 52.14 {\color{red} (-0.98)} & 79.72 {\color{red} (-0.31)} \\
 & W8A8  & 71.08 {\color{red} (-0.34)} & 75.82 {\color{red} (-0.07)} & 67.14 {\color{green} (+0.03)} & 83.15 {\color{red} (-0.13)} & 50.55 {\color{red} (-0.61)} & 80.43 {\color{green} (+0.4)} \\
 \hline
 \multirow{3}{*}{\textbf{Llama-7B}} & FP16  & 49.83  & 23.2 & 59.34 & 47.22 & 45.04 & 72.93 \\
 & AWQ   & 48.98 {\color{red} (-0.85)}& 21.23{\color{red} (-1.97)} & 58.61 {\color{red} (-0.73)} & 45.34 {\color{red} (-1.88)} & 43.57 {\color{red} (-1.47)} & 72.53 {\color{red} (-0.4)} \\
 & W8A8  & 50.34 {\color{green} (+0.51)} & 22.67 {\color{red} (-0.53)} & 59.3 {\color{red} (-0.04)} & 47.24 {\color{green} (+0.02)} & 44.80 {\color{red} (-0.24)} & 73.32 {\color{green} (+0.39)} \\
\hline
 \multirow{3}{*}{\textbf{Llama-13B}} & FP16  & 55.63  & 35.56 & 63.1 & 53.55 & 40.88 & 75.06 \\
 & AWQ   & 54.95 {\color{red} (-0.68)}& 31.69{\color{red} (-3.87)} & 62.13 {\color{red} (-0.97)} & 53.77 {\color{green} (+0.22)} & 41.37 {\color{green} (+0.49)} & 76.09 {\color{green} (+1.03)} \\
 & W8A8  & 55.29 {\color{red} (-0.34)} & 35.18 {\color{red} (-0.38)} & 63.08 {\color{red} (-0.02)} & 53.65 {\color{green} (+0.1)} & 41.49 {\color{green} (+0.61)} & 75.22 {\color{green} (+0.16)} \\
 \bottomrule
\end{tabular}
\end{adjustbox}
\end{table*}

As shown in \Cref{tab:benchmark_quant}, we used LM Eval~\cite{eval-harness}, an open-source LLM evaluation tool, to assess the LLMs used in our experiments and their quantized versions. The evaluations were conducted on tasks from the Open LLM Leaderboard, including ARC-c~\cite{clark2018arc}, GSM8k~\cite{cobbe2021gsm8k}, HellaSwag~\cite{zellers2019hellaswag}, MMLU~\cite{hendrycks2020mmlu}, TruthfulQA~\cite{lin2021truthfulqa}, and Winogrande~\cite{sakaguchi2021winogrande}.

We also present the Qscore for the LLMs and their quantized versions across three datasets, as shown in \Cref{tab:quant_qscore}.

\begin{table}[!t]
\centering
\caption{Mean Qscore on three datasets for Qwen and Llama families with their quantized versions.}
\label{tab:quant_qscore}
\begin{adjustbox}{width=0.48\textwidth}
\begin{tabular}{c|c|c|c|c}
\toprule
\textbf{Model} & \textbf{Method}  & \makecell{\textbf{Qscore}\\\footnotesize{NewsQA}} & \makecell{\textbf{Qscore}\\\footnotesize{ArenaHard}} & \makecell{\textbf{Qscore}\\\footnotesize{HumanEval}}  \\
\midrule
\multirow{3}{*}{\textbf{Qwen-7B}} & FP16  & 11.11 & 15.90 & 26.82  \\
 & AWQ  & 11.77 {\color{green} (+0.66)} & 14.17 {\color{red} (-1.73)} & 26.52 {\color{red} (-0.3)}  \\
 & W8A8  & 10.46 {\color{red} (-0.65)} & 15.76 {\color{red} (-0.14)} & 26.84 {\color{green} (+0.02)} \\
\hline
\multirow{3}{*}{\textbf{Qwen-14B}} & FP16  & 14.37 & 18.41 & 27.03  \\
 & AWQ  & 12.01 {\color{red} (-2.36)} & 15.33 {\color{red} (-3.08)} & 26.41 {\color{red} (-0.62)}  \\
 & W8A8  & 14.40 {\color{green} (+0.03)} & 18.49 {\color{green} (+0.08)} & 27.24 {\color{green} (+0.21)} \\
\hline
\multirow{3}{*}{\textbf{Qwen-32B}} & FP16  & 15.62 & 19.82 & 28.86  \\
 & AWQ  & 14.87 {\color{red} (-0.75)} & 18.89 {\color{red} (-0.93)} & 27.99 {\color{red} (-0.87)}  \\
 & W8A8  & 15.36 {\color{red} (-0.26)} & 19.73 {\color{red} (-0.09)} & 28.38 {\color{red} (-0.48)} \\
\hline
\multirow{3}{*}{\textbf{Llama-7B}} & FP16  & 6.81 & -7.93 & -4.19  \\
 & AWQ  & 6.49 {\color{red} (-0.32)} & -10.19 {\color{red} (-2.26)} & -8.23 {\color{red} (-4.04)}  \\
 & W8A8  & 6.87 {\color{green} (+0.06)} & -8.58 {\color{red} (-0.65)} & -4.23 {\color{red} (-0.04)} \\
\hline
\multirow{3}{*}{\textbf{Llama-13B}} & FP16  & 8.73 & -5.63 & 0.41  \\
 & AWQ  & 8.63 {\color{red} (-0.11)} & -6.34 {\color{red} (-0.71)} & -1.65 {\color{red} (-2.06)}  \\
 & W8A8  & 8.64 {\color{red} (-0.09)} & -5.69 {\color{red} (-0.06)} & 0.24 {\color{red} (-0.17)} \\
\bottomrule
\end{tabular}
\end{adjustbox}
\end{table}

\subsection{Results on NewsQA Summarization}
\begin{figure}[!t]
    \centering
    \includegraphics[width=0.47\textwidth]{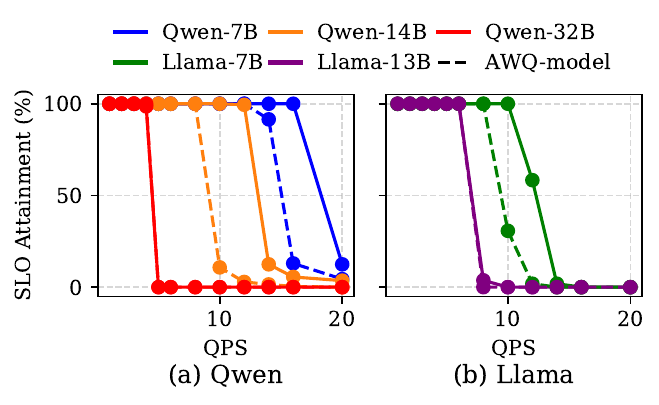}
    \caption{SLO attainment of Qwen and Llama families with AWQ version across QPS range.}
    \label{fig:quant_slo_awq}
\end{figure}

\begin{figure}[!t]
    \centering
    \includegraphics[width=0.47\textwidth]{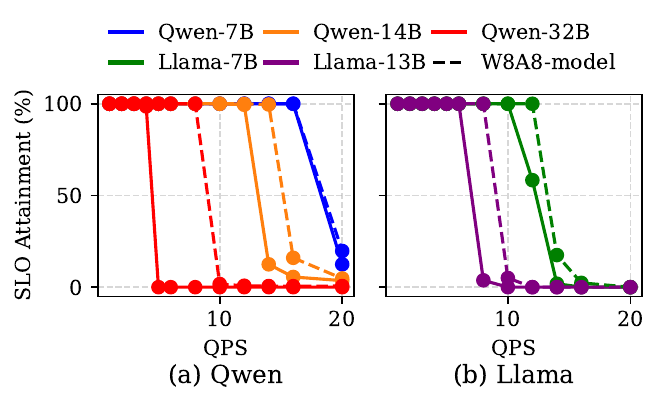}
    \caption{SLO attainment of Qwen and Llama families with W8A8 version across QPS range.}
    \label{fig:quant_slo_w8a8}
\end{figure}

\Cref{fig:quant_slo_awq} and \Cref{fig:quant_slo_w8a8} illustrate the impact of the AWQ and W8A8 quantized versions on SLO attainment across the QPS range for the Qwen and Llama families. As shown in Figure \ref{fig:quant_slo_awq}, the AWQ version of the models fails to meet the SLO at lower QPS values. In contrast, the W8A8 quantized version improves efficiency, enabling models to serve a higher QPS.

\begin{figure}[!t]
    \centering
    \includegraphics[width=0.48\textwidth]{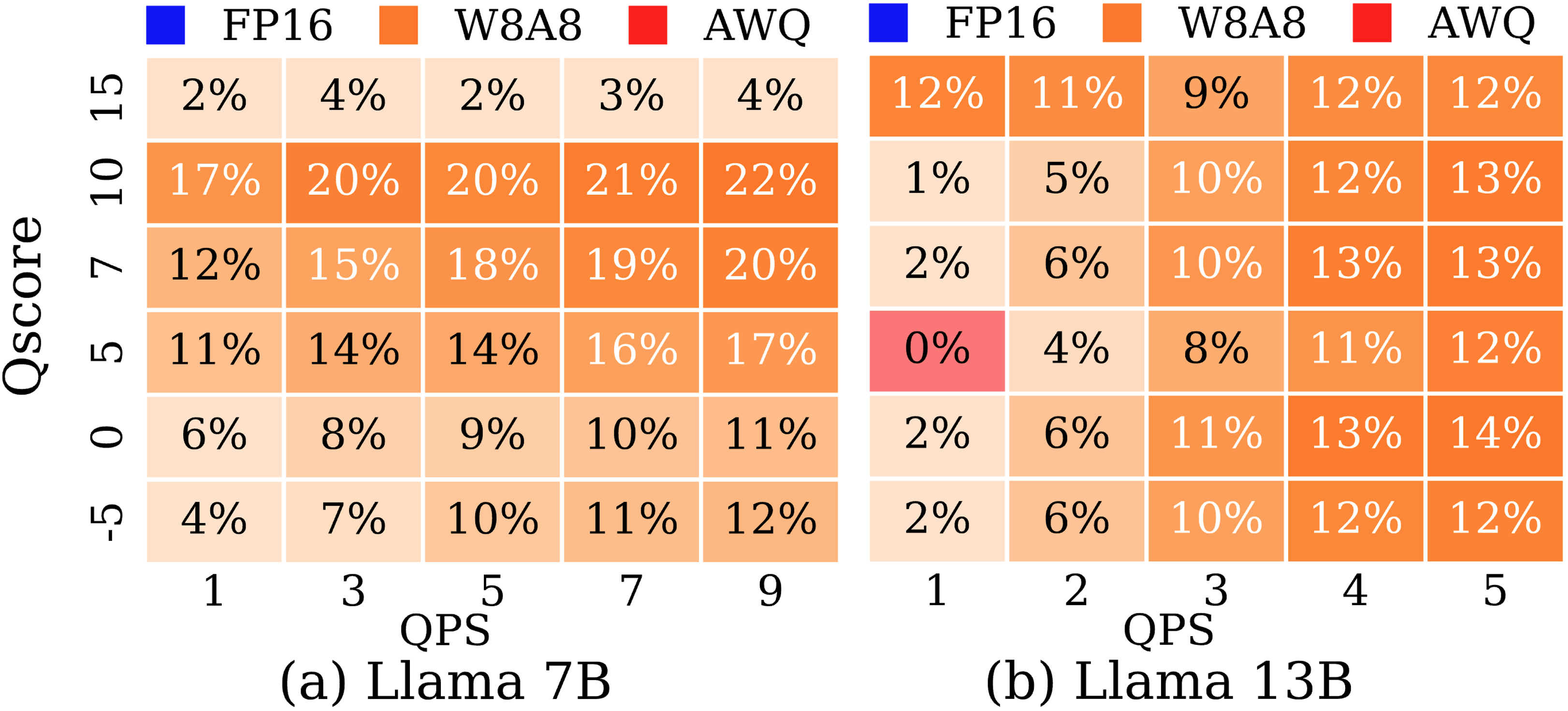}
    \caption{Comparison of FP16, AWQ and W8A8 versions of Llama 7B/13B in \SYSTEM{} on NewsQA dataset. Tile colors indicate the model with the lowest carbon per FU. Tile values are carbon savings (\%) of greenest version compared to the second greenest.}
    \label{fig:quant_newsqa_llama_heatmap}
\end{figure}

\Cref{fig:quant_newsqa_llama_heatmap} shows the comparison results among FP16, AWQ and W8A8 versions in Llama family. W8A8 has almost the lowest carbon emission under all conditions.

\subsection{Results on Arena Hard}
\begin{figure}[!t]
    \centering
    \includegraphics[width=0.48\textwidth]{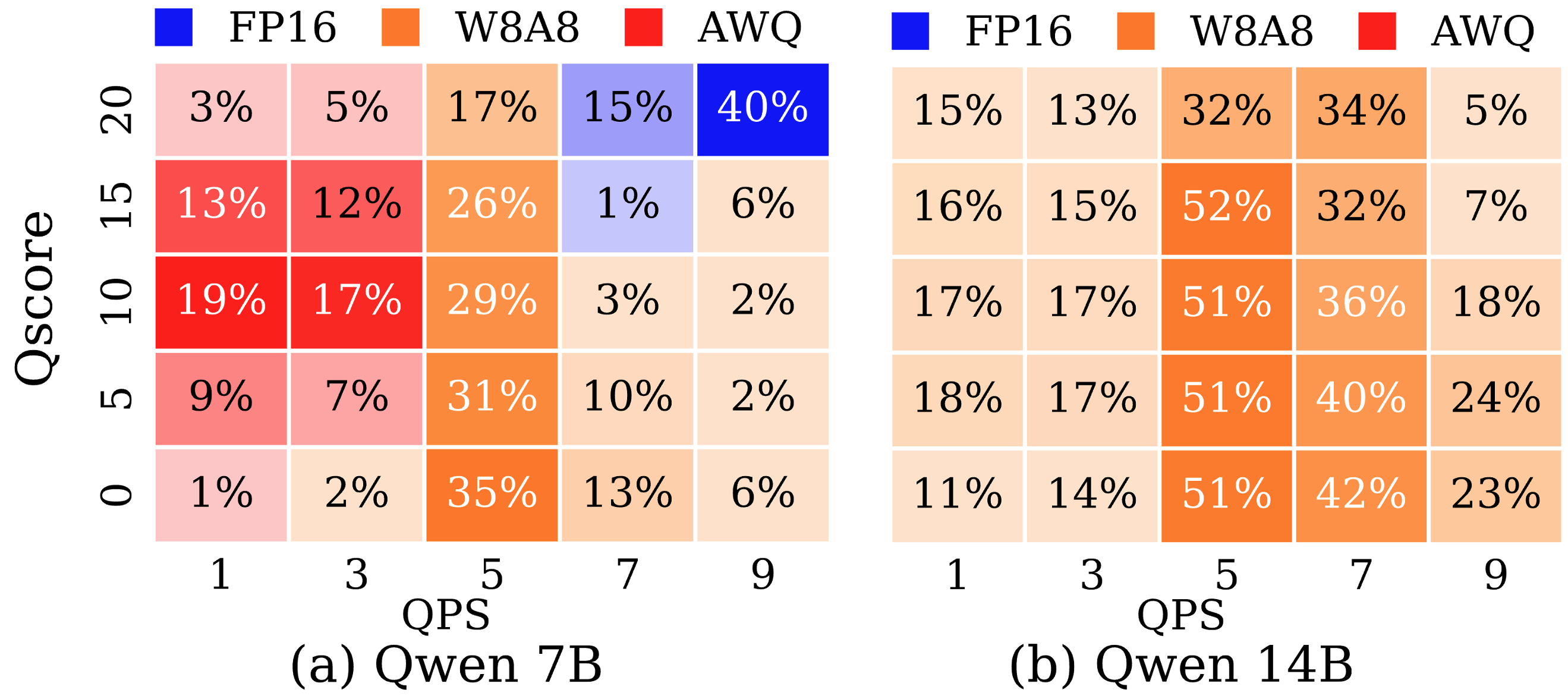}
    \vspace{-0.2in}
    \caption{Comparison of FP16, AWQ and W8A8 versions of Qwen 7B/14B in \SYSTEM{} on Arena Hard dataset. Tile colors indicate the model with the lowest carbon per FU. Tile values are carbon savings (\%) of the greenest version compared to the second greenest.}
    \label{fig:arena_quant_heat_w8a8_qwen}
\end{figure}

\begin{figure}[!t]
    \centering
    \includegraphics[width=0.48\textwidth]{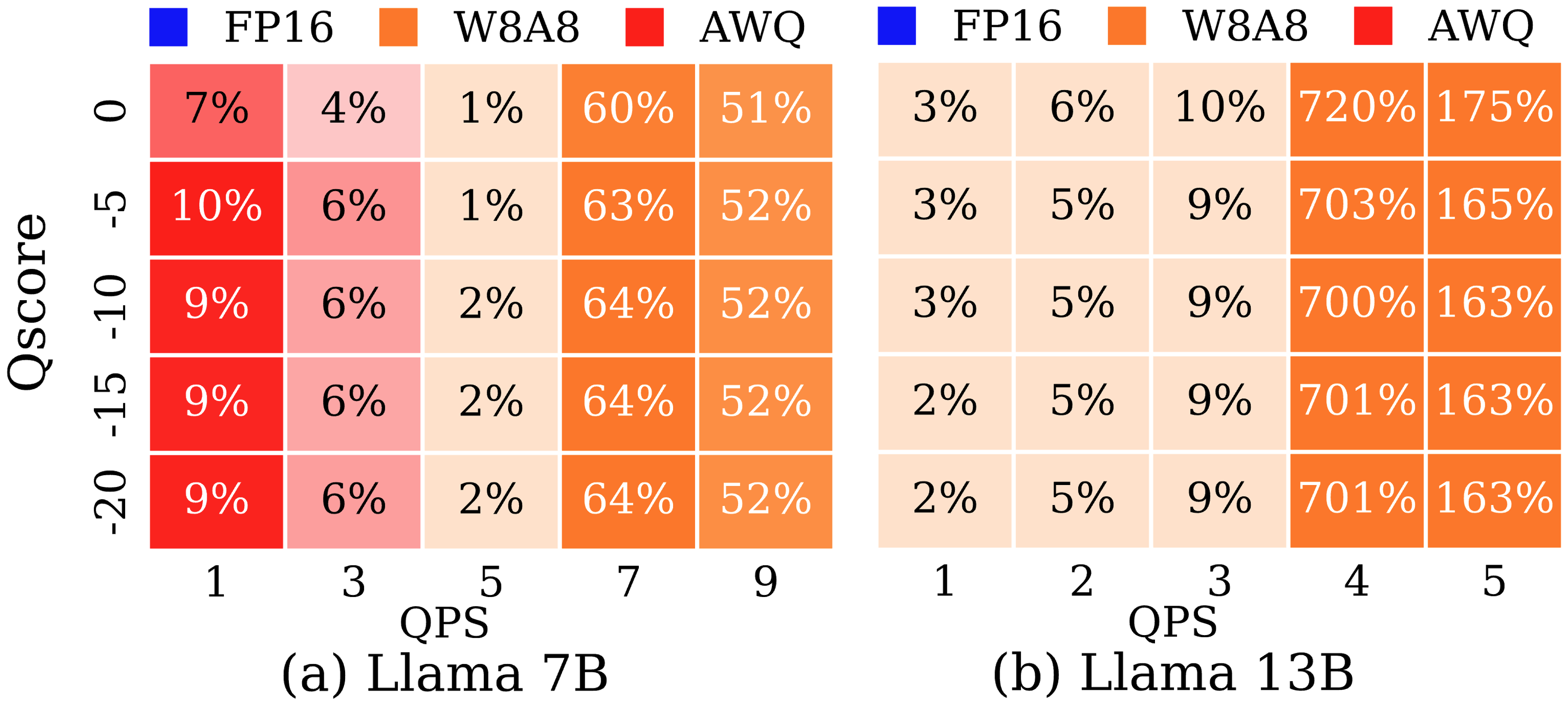}
    \vspace{-0.2in}
    \caption{Comparison of FP16, AWQ and W8A8 versions of Llama 7B/13B in \SYSTEM{} on Areana Hard dataset. Tile colors indicate the model with the lowest carbon per FU. Tile values are carbon savings (\%) of greenest version compared to the second greenest.}
    \label{fig:arena_quant_heat_w8a8_llama}
\end{figure}

\Cref{fig:arena_quant_heat_w8a8_qwen} shows the results of different quantization methods for Qwen 7B/14B models on Arena Hard dataset. This aligns well with the results on the previous NewsQA summarization dataset, as AWQ shows an advantage at low QPS on the smaller 7B model. When we use 14B model, W8A8 illustrates the great potential to save up to 50\% carbon emission under each scenario. We can see a similar trend in \Cref{fig:arena_quant_heat_w8a8_llama} on Llama models. 
 
\subsection{Results on HumanEval}
\begin{figure}[!t]
    \centering
    \includegraphics[width=0.48\textwidth]{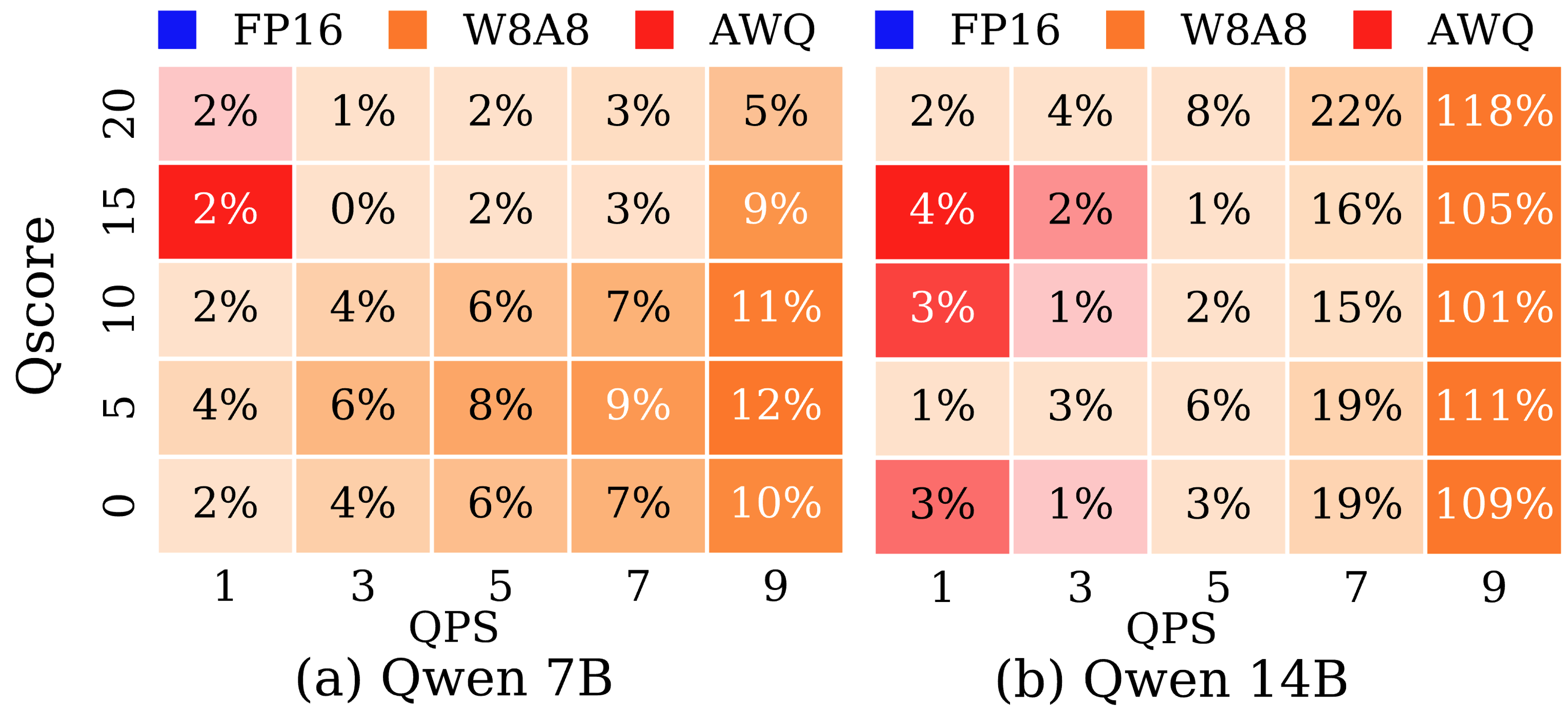}
    \vspace{-0.2in}
    \caption{Comparison of FP16, AWQ and W8A8 versions of Qwen 7B/14B in \SYSTEM{} on HumanEval dataset. Tile colors indicate the model with the lowest carbon per FU. Tile values are carbon savings (\%) of greenest quantization version compared to the second greenest.}
    \label{fig:humaneval_quant_heat_w8a8_qwen}
\end{figure}

As shown in \Cref{fig:humaneval_quant_heat_w8a8_qwen}, AWQ still becomes the greenest method when QPS is low, but W8A8 dominates in more conditions on the Qwen 7B model. This is because, after W8A8 quantization, the Qscore of Qwen 7B improves on the HumanEval dataset. For the Qwen 14B model, W8A8 is no longer the greenest method under all conditions. This is due to AWQ experiencing minimal accuracy degradation on this dataset, allowing it to retain its advantage at low QPS.

\begin{figure}[!t]
    \centering
    \includegraphics[width=0.48\textwidth]{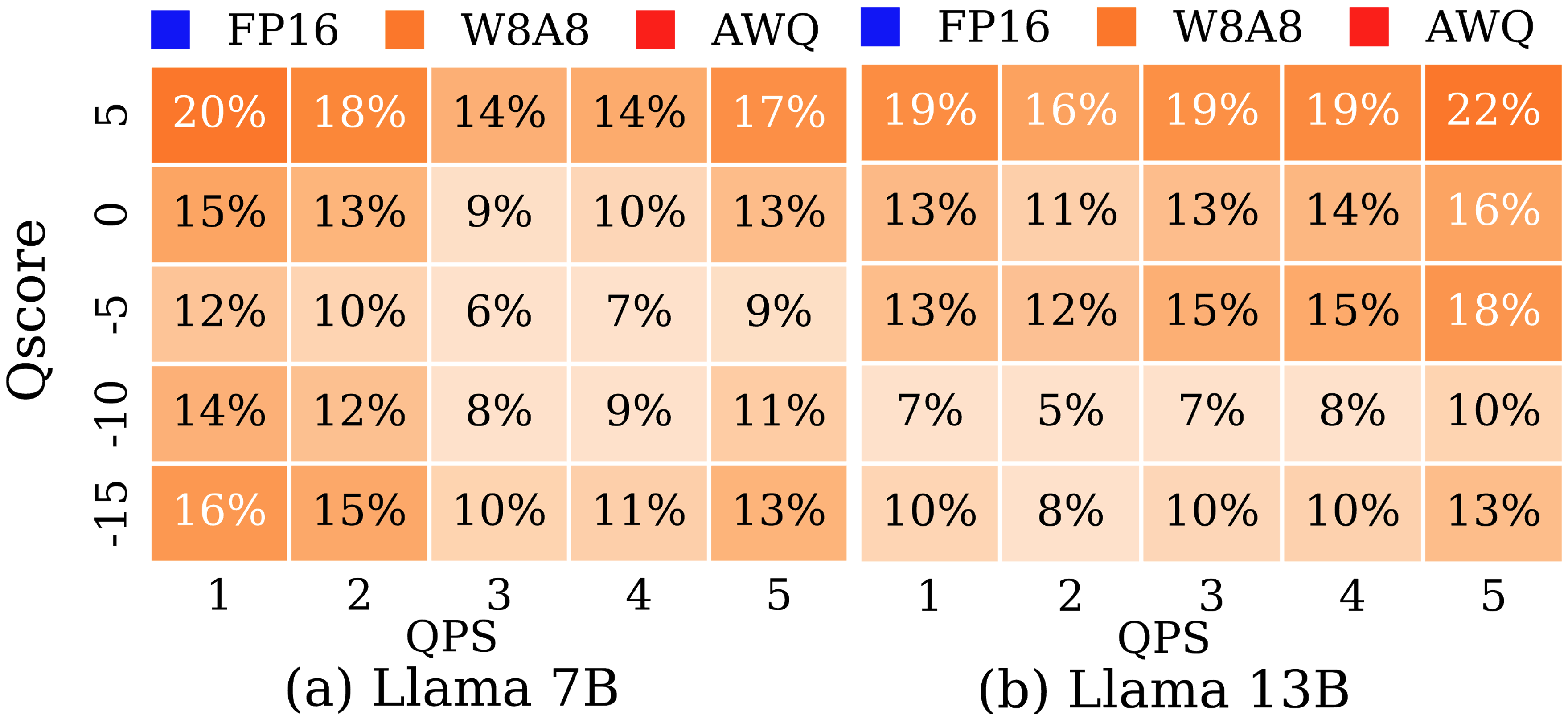}
    \vspace{-0.2in}
    \caption{Comparison of FP16, AWQ and W8A8 versions of Llama 7B/13B in \SYSTEM{} on HumanEval dataset. Tile colors indicate the model with the lowest carbon per FU. Tile values are carbon savings (\%) of greenest quantization version compared to the second greenest.}
    \label{fig:humaneval_quant_heat_w8a8_llama}
\end{figure}

\begin{figure}[!t]
    \centering
    \includegraphics[width=0.48\textwidth]{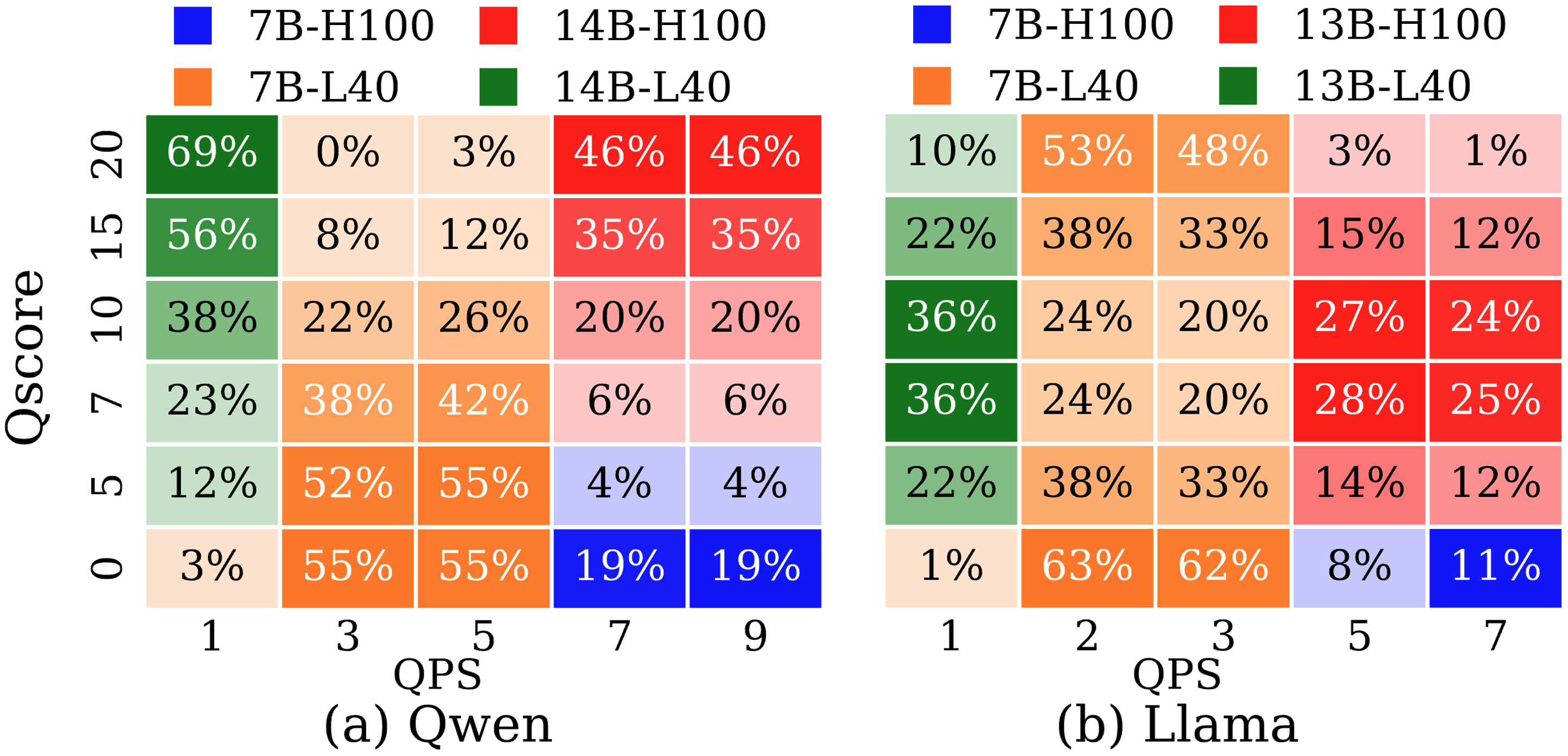}
    \caption{Comparison of model-hardware combinations for Qwen and Llama in \SYSTEM{}. Tile colors indicate the model-hardware with the lowest carbon per FU. Tile values are carbon savings (\%) of the greenest choice compared to the second greenest.}
    \label{fig:hw_heatmap_all}
\end{figure}

\section{Additional Results for Hardware Case Study}
\label{sec:appendix_hardware}
\Cref{fig:hw_heatmap_all} compares model and hardware combinations, further confirming our previous conclusion: older hardware can achieve lower carbon emissions under certain conditions. As shown in the figure, whether for the Qwen or Llama model families, the greenest choice at low QPS is consistently the L40 server. Once the hardware is fixed, we can apply insights from the model size case study to select the model size based on the quality requirement. This pattern reaffirms that choosing the optimal model and hardware combination requires a balance between performance needs and carbon efficiency.

\end{document}